\documentclass[runningheads]{llncs}

\usepackage{eccv}

\usepackage{eccvabbrv}

\usepackage{graphicx}
\usepackage{booktabs}
\usepackage{multirow}
\usepackage{colortbl}
\usepackage{tcolorbox}
\usepackage[accsupp]{axessibility}  

\usepackage{hyperref}

\begin{document}

\title{LACO: Adaptive Latent Communication for Collaborative Driving} 

\titlerunning{LACO}

\author{Tianhao Chen\and
Yuheng Wu \and
Dongman Lee}
\authorrunning{T.Chen et al.}

\institute{
Korea Advanced Institute of Science \& Technology \\ 
\email{\{thchen,yuhengwu,dlee\}@kaist.ac.kr}
}
\maketitle
\begin{abstract}
Collaborative driving aims to improve safety and efficiency by enabling connected vehicles to coordinate under partial observability. Recent approaches have evolved from sharing visual features for perception to exchanging language-based reasoning through foundation models for behavioral coordination. Though communicating in language provides intuitive information, it introduces two challenges: high latency caused by autoregressive decoding and information loss caused by compressing rich internal representations into discrete tokens.
To address these challenges, we analyze latent communication in collaborative driving under inherent limitations of multi-agent settings. Our analysis reveals agent identity confusion, where direct fusion of latent states entangles decision representations across vehicles. Motivated by this, we propose LACO, a training-free \textbf{LA}tent \textbf{CO}mmunication paradigm that seamlessly adapts pretrained driving models to collaborative settings. LACO introduces Iterative Latent Deliberation (ILD) for latent reasoning, Cross-Horizon Saliency Attribution (CHSA) for communication-efficient information selection, and Structured Semantic Knowledge Distillation (SSKD) to stabilize ego-centric decision making. Closed-loop experiments in CARLA show that LACO notably reduces communication and inference latency while maintaining strong collaborative driving performance.
\end{abstract}
\section{Introduction}
\label{sec:intro}
\begin{figure*}[t]
  \centering
  \includegraphics[width=\textwidth]{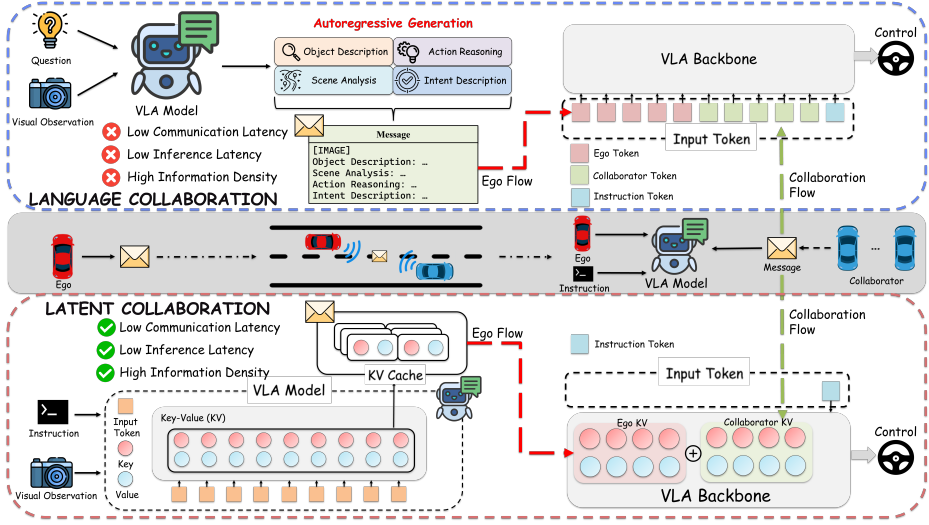}
    \caption{\textbf{Communication interfaces for multi-agent VLA collaboration.} 
    \textbf{Top:} Language-based collaboration depends on explicit information gathering and multi-round natural language exchange, leading to latency and information loss. 
    \textbf{Bottom:} Latent collaboration directly exchanges internal latent representations (KV caches) during inference  removing the linguistic bottleneck and achieve effectiveness.}
  \label{fig:concept}
\end{figure*}
Collaborative driving seeks to improve safety and traffic efficiency by enabling connected vehicles to coordinate perception, intention, and action under partial observability \cite{han2023collaborative}. Early collaborative systems \cite{wu2025background,gao2025stamp,wang2020v2vnet,xu2022v2x,xu2025cosdh,you2026v2x} predominantly focused on sharing perception, \eg visual embeddings or intermediate features, to expand each agent’s field of view. While effective for mitigating sensing limitations, perception-only exchange overlooks that safe coordination also depends on reasoning about the scene, such as inferring intent, assessing risk, and planning multi-step interactions. With the emergence of foundation models equipped with strong reasoning capabilities, language-based collaboration \cite{gao2025langcoop,chiu2025v2v,wu2025v2x,jiang2024koma} has become a dominant paradigm: agents communicate reasoning results in natural language, including beliefs, intentions, and planned maneuvers, to resolve uncertainty and coordinate actions.

Despite its success, language-based communication introduces practical limitations for collaborative driving. First, it requires an explicit information collection stage and an autoregressive \cite{vaswani2017attention} decoding process, which jointly increase computation and end-to-end latency and are ill-suited for real-time decision making. Second, language is a lossy compression channel: translating high dimensional visual evidence and latent reasoning into a short token sequence can omit critical coordination details and degrade performance.

Overcoming the above two challenges is non-trivial. Inspired by prior works on latent reasoning and KV-based communication \cite{zou2025latent,dery2026latent,du2025enabling,fu2025cache,ye2025kvcomm}, we argue that a more effective interface for collaboration lies in the latent space, where the internal representations that actually drive decision making reside. During inference, task-relevant information is already encoded in the model’s hidden states, without the need to decode it into a low-dimensional linguistic form. This observation suggests a latent communication paradigm that avoids the linguistic bottleneck altogether.
Therefore, for collaborative driving, a more faithful communication channel is to directly share the model’s \emph{thinking state}. Transformer-based agents \cite{vaswani2017attention} maintain rich intermediate representations throughout inference; in particular, the key–value (KV) cache provides a structured record of perception and ongoing reasoning. Building on this property, we propose a training-free latent collaboration paradigm for autonomous driving. Instead of relying on multi-round language exchanges, vehicles directly transmit selected KV caches generated during inference. This design bypasses token-level communication and eliminates additional autoregressive decoding, enabling higher-fidelity information sharing with significantly reduced communication and inference latency.

However, deploying latent communication in collaborative driving poses challenges that differ structurally from prior multi-agent settings. Existing latent-collaboration frameworks typically decompose a task into multiple sub-problems, where agents specialize in different stages or components and exchange intermediate representations to facilitate coordination or reuse computation \cite{guo2024large,zhang2024chain,hong2023metagpt}. Collaborative driving, in contrast, is inherently parallel: each vehicle must complete its own inference, and the received information is not subsumed by its local computation, but instead provides an additional perspective. This creates an architectural mismatch where task-decomposed information exchange contrasts with parallel, ego-centric inference, and shifts the communication objective from intermediate reuse to perspective augmentation under a strict computational constraint. Consequently, conventional multi-agent latent schemes are not directly compatible with collaborative driving without rethinking how and where latent information is integrated.Through extensive experiments, we find that naive full-KV fusion \cite{zou2025latent,ye2025kvcomm} can induce \emph{agent identity confusion}, where the receiver over-attends to another agent’s internal states, leading to cross-agent representation entanglement. Our framework addresses these challenges in three complementary steps: Iterative Latent Deliberation (ILD) embeds decision knowledge into latent states with minimal additional cost; Cross-Horizon Saliency Attribution (CHSA) then compresses transmission by selecting only reasoning-critical information from reasoning traces; and Shallow-Stream Knowledge Distillation (SSKD), designed via our attention-pattern analysis, reduces bandwidth and mitigates KV confusion by transmitting shallow-layer KV to provide global context while preserving ego-centric representation integrity. These in all yield a communication interface that is both informative for coordination and stable for downstream decision making.

\noindent\textbf{Contribution}. Our contributions are threefold: (1) We introduce latent communication for collaborative driving by exchanging transformer KV caches instead of language-based information collection. (2) We propose communication-aware KV transmission that integrates latent reasoning for selective sharing with a layer-wise fusion scheme for robustness. (3) We conduct closed-loop evaluations with CARLA \cite{dosovitskiy2017carla}, showing that our approach significantly reduces communication and inference latency while preserving strong collaborative driving performance.
\section{Related Work}
\subsection{Language Collaboration for Driving}
Traditional collaborative driving approaches improve perception and situational awareness by exchanging information at different stages: (i) early-fusion methods share raw sensor data such as point clouds \cite{chen2019cooper,arnold2020cooperative}, (ii) intermediate-fusion approaches transmit compact representations like BEV features \cite{chiu2025v2v,gao2025stamp,liu2020when2com,hu2022where2comm,wu2025background}, and (iii) late-fusion methods communicate high-level outputs such as detected bounding boxes~\cite{xu2022opv2v,xu2025cosdh}. While effective for enhancing perception, these pipelines do not explicitly coordinate intent among agents, which limits the intelligence and adaptability of collaborative driving. 

Recent language-based collaboration methods leverage reasoning ability of foundation models to negotiate intent, assuming that the underlying models (\eg, VLMs) can be jointly trained or fine-tuned to support inter-agent communication \cite{gao2025langcoop,chiu2025v2v,wu2025v2x,jiang2024koma}. However, real-world autonomous driving systems predominantly deploy ego-centric VLA models \cite{fu2025orion,renz2025simlingo,jiang2025survey,shao2024lmdrive,hwang2024emma,sima2024drivelm}. Retraining these large pre-trained models for collaboration is resource-intensive and risks degrading the pre-existing capabilities of the VLA. Moreover, language-based coordination requires generative capabilities, which many VLA models \cite{yang2025drivemoe, wang2025alpamayo, li2025recogdrive, karypidis2024dino} inherently lack due to their training paradigm and architectural design. Finally, autoregressive process for language generation introduces high latency and the decoding process from hidden states to language brings information loss, making such approaches unsuitable for real-time and safety-critical driving scenarios.

Our work instead focuses on enabling robust collaboration directly among ego-centric VLA models, achieving low-latency, intent-aware coordination without relying on explicit language generation or retraining.
\subsection{Multi-agent Latent Collaboration}
Recent works in latent collaboration have shown that exchanging internal representations across agents enables efficient multi-agent reasoning \cite{zou2025latent,dery2026latent,du2025enabling,fu2025cache,ye2025kvcomm}, which demonstrate that latent-space communication can bypass text-based bottlenecks while preserving rich information. For instance, \cite{dery2026latent} fosters communications across models via a shared KV‑cache latent space without changing model parameters. \cite{ye2025kvcomm} aligns and reuses agents’ KV caches via an online anchor pool, enabling efficient latent context sharing. However, extending these ideas to physical-world autonomous driving is non-trivial due to strict bandwidth limits and the ego-centric nature of the task: unlike multi-agent language tasks \cite{guo2024large,zhang2024chain,hong2023metagpt}, there is no natural sub-task decomposition in collaborative driving, and all agents act independently in overlapping environments. To the best of our knowledge, we take the first step toward latent collaboration between ego-centric VLA agents, proposing a mechanism that allows multiple autonomous vehicles to exchange compressed latent representations while respecting real-world constraints.

\begin{figure*}[t]
  \centering
  \includegraphics[width=\textwidth]{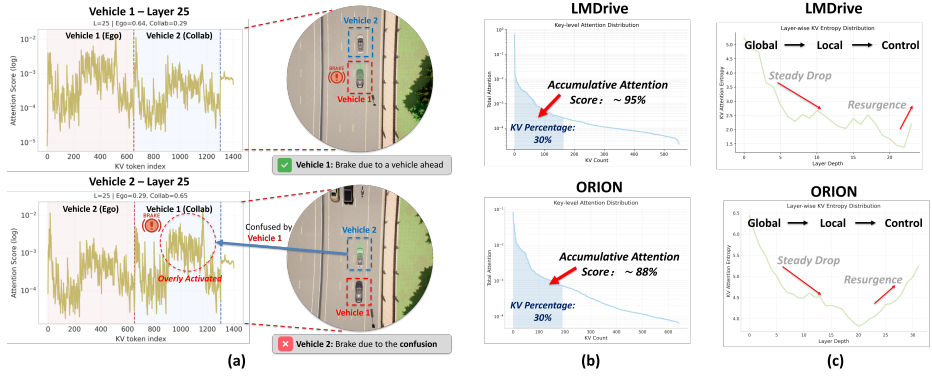}
    \caption{\textbf{(a) Agent Identity Confusion:} Under full-depth fusion, the ego vehicle over-attends to collaborator latent states, resulting in policy hijacking and erroneous maneuvers. 
    \textbf{(b) Attention Sparsity:} A small fraction ($\approx$ 30\%) of tokens captures the majority of attention mass, revealing substantial spatial redundancy.
    \textbf{(c) Attention Entropy Analysis:} Layer-wise entropy exhibits a global-to-local transition, followed by a late-stage resurgence during control synthesis.}
  \label{fig:motivation}
\end{figure*}

\section{Motivation Study}
\label{sec:motivation}

Following \cite{zou2025latent,ye2025kvcomm}, a straightforward strategy for collaborative VLA driving is to transmit the full reasoning KV cache between agents. However, we observe a critical instability: \textit{the ego agent’s control policy can be unexpectedly biased or even overridden by collaborator states, despite contradictory local observations.}

As shown in \cref{fig:motivation}(a), the ego vehicle may brake due to a hazard visible only from a collaborator’s viewpoint, even when its own lane is clear. Attention visualization reveals that deep-layer activations become disproportionately amplified toward collaborator tokens under full-depth fusion. Since late-stage representations are tightly coupled with control synthesis, this amplification directly perturbs ego-centric decision formation.
We term this phenomenon \emph{agent identity confusion}. It suggests that naive full-depth latent fusion does not merely aggregate complementary perception, but can disrupt identity-consistent control.
To understand this instability, we analyze VLA representations along two axes: spatial attention distribution and depth-wise representational dynamics.
\newcounter{observation}[section] 
\refstepcounter{observation} 

\noindent\textbf{Observation \arabic{observation}: Structured Sparsity in Visual Reasoning}
\label{ob1}
Aggregating attention weights across layers and heads reveals a pronounced long-tail distribution (\cref{fig:motivation}(b)): a small subset of visual tokens captures most attention mass, while many remain weakly activated.
This indicates that spatial reasoning is intrinsically sparse. The model progressively contracts attention onto decision-relevant anchors that dominate control formation.
From a communication perspective, the prefill \cite{kwon2023efficient} KV cache is dominated by high-resolution visual tokens. Under full transmission, bandwidth scales with visual resolution rather than decision contribution. The long-tail structure therefore implies a structural inefficiency: transmitting all tokens expands communication cost without proportionally increasing effective decision information, and may introduce additional representational variability into later layers.

\noindent\textbf{Insight 1.} Full visual KV transmission propagates a large set of low-impact representations, inflating bandwidth and potentially destabilizing downstream decision consolidation.

\refstepcounter{observation} 
\noindent\textbf{Observation \arabic{observation}: Agent Identity Confusion in Multi-Agent Collaboration}%
\label{ob2}
Spatial sparsity alone does not explain agent identity confusion. We therefore analyze depth-wise representational dynamics under full-layer KV fusion.
We compute layer-wise attention entropy (\cref{fig:motivation}(c)) as
\begin{equation}
e^{(l)} = - \frac{1}{H} \sum_{h=1}^{H} \sum_{j=1}^{N}
\alpha_{h,j}^{(l)} \log (\alpha_{h,j}^{(l)} + \epsilon)
\label{eq:layer_attention_entropy}
\end{equation}
Entropy follows a consistent U-shaped trajectory: early layers exhibit high entropy corresponding to global scene parsing; intermediate layers show entropy collapse as attention contracts onto ego-conditioned evidence; deep layers display entropy resurgence during action synthesis.
This pattern reflects a depth-wise transition from global perception to ego-centric decision structuring, and finally to tightly coupled control synthesis. At deep layers, perception and emerging intent become geometrically entangled within a control-oriented space.
Under full-depth fusion, collaborator KV states are injected precisely into this entangled regime. The cache at this stage encodes not only perception but perspective-conditioned interpretation and partially synthesized control tendencies. Mixing such representations across agents introduces competing policy structures into a control-sensitive space, creating structural conditions for perceptual inconsistency and intent-level interference.

\noindent\textbf{Insight 2.}
Agent confusion emerges when fusion occurs after perception has become entangled with agent-specific control synthesis. In contrast, shallow representations remain globally informed yet structurally disentangled, making early-stage exchange a more stable interface for collaboration.

\noindent\textbf{Implications}
Collaborative instability arises from two structural factors: spatial over-transmission and depth-wise fusion of identity-entangled representations. Effective communication must therefore be (1) saliency-aware, prioritizing structurally significant tokens, and (2) depth-aware, occurring before perception becomes inseparable from agent-specific control synthesis.
These principles motivate a new latent collaboration framework for collaborative driving.
\section{Method}
\label{sec:methodology}
\subsection{Overview}
\begin{figure}[t]
    \centering
    \includegraphics[width=\linewidth]{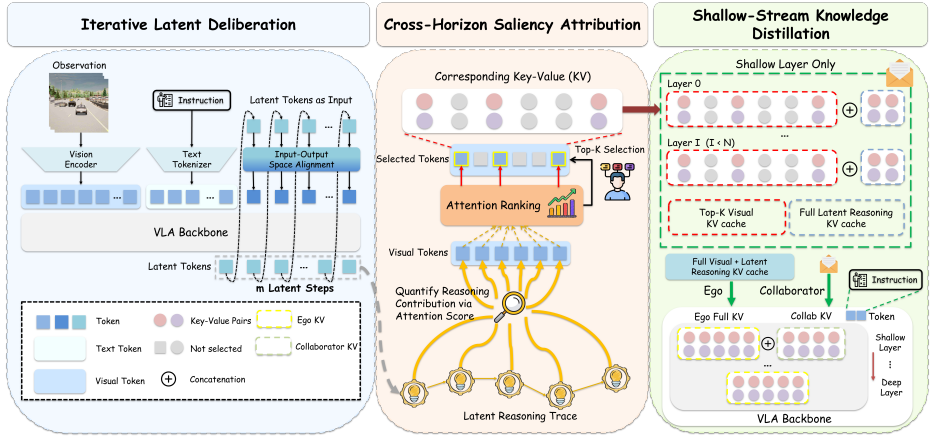}
    \caption{
    Overview of LACO. Each agent performs Iterative Latent Deliberation (ILD) to embed spatial state and intent into its KV cache, selects decision-critical tokens via Cross-Horizon Saliency Attribution (CHSA), and transmits only early-layer representations through Shallow-Stream Knowledge Distillation (SSKD). The receiver integrates the message with its own reasoning stack for collaborated inference.
    }
    \label{fig:framework}
\end{figure}

We propose a communication-efficient collaborative framework that restructures the pipeline of perception and coordination in VLA driving as shown in \cref{fig:framework}. Instead of transmitting full internal states, each vehicle first consolidates its perception and emerging intent through latent reasoning, ensuring that communication originates from semantically organized representations rather than raw visual tokens.
The internal state is then selectively compressed according to the model’s intrinsic saliency structure, reducing the long tail of weakly contributing representations and controlling payload growth. Crucially, communication is restricted to early-stage representations, before perceptual abstractions become tightly coupled with control synthesis. By exchanging shallow-layer states instead of deeply entangled decision structures, the framework mitigates cross-agent identity interference while preserving globally informative context.
The receiver integrates this compressed context into its own reasoning process, enabling coordination without destabilizing ego-centric control formation.
\label{subsec:overview}
\subsection{Iterative Latent Deliberation (ILD)}
\label{subsec:ild}
Relying on natural language to articulate complex beliefs and intentions introduces a significant lossy compression channel and an autoregressive latency bottleneck. Previous works \cite{saunshi2025reasoning,hao2024training,zhu2025scaling,tan2025think} in latent reasoning have demonstrated that models can express their internal thinking states more effectively by leveraging continuous hidden representations rather than discrete tokens. This approach allows the model to preserve nuanced semantic structures and multi-modal uncertainties that are typically discarded during the language decoding process.
To leverage these latent advantages within the real-time constraints of collaborative driving, we propose \textbf{Iterative Latent Deliberation (ILD)}, which operates entirely ego-centrically, iteratively embedding the vehicle's spatial reasoning, risk assessment, and decision-making rationale directly into its internal Key-Value (KV) cache with minimal computational overhead.

\noindent\textbf{Latent Reasoning.}
During initial inference, the ego vehicle processes input embeddings $E = [e_1, \dots, e_T]$ through $L$ transformer layers to obtain the final-layer hidden state $h^{(0)}$ and populate the initial KV cache $\mathcal{K}\mathcal{V}{prefill}$. Following \cite{zou2025latent}, rather than projecting $h^{(0)}$ to language space, reasoning unfolds entirely in latent space via $m$ iterative forward passes, where at each step $t \in [1, m]$, $h^{(t-1)}$ is fed back to produce $h^{(t)}$ and append new keys and values to the cache. Let $W_{in} \in \mathbb{R}^{|V| \times d}$ and $W_{out} \in \mathbb{R}^{d \times |V|}$ be the input embedding matrix and the output language head, respectively. To mitigate out-of-distribution activations during recursion, a lightweight linear projection aligns the hidden states:
\begin{equation} 
\hat{e}^{(t)} = h^{(t-1)} W_a \quad \text{where} \quad W_a \approx W_{out}^{\dagger} W_{in} 
\end{equation} 
where $W_{out}^{\dagger}$ is the pseudo-inverse of $W_{out}$.
The projection is computed once and reused, introducing negligible computational overhead.

\noindent\textbf{Dual-Purpose Latent Trace Preparation.}
Repeating the latent update for a fixed number of steps yields a compact latent trace and its reasoning KV cache. Crucially, this fixed-step deliberation serves a dual structural purpose. First, it seamlessly prefills the visual and contextual observation KV cache ($\mathcal{K}\mathcal{V}_{prefill}$) for the ego vehicle, effectively bypassing redundant feature extraction during the final decision-making stage. Second, within a strictly controllable computational budget ($m$ steps), it progressively embeds the vehicle's high level decision-making rationale, spatial semantics, and prospective driving intent directly into the latent representations.
\subsection{Cross-Horizon Saliency Attribution (CHSA)}
\label{subsec:chsa}
Transmitting the full reasoning trace ($\mathcal{K}\mathcal{V}{prefill} \cup \mathcal{K}\mathcal{V}{latent}$) to neighboring vehicles is infeasible under real-time bandwidth constraints. Motivated by Observation~\ref{ob1}, we propose \textbf{Cross-Horizon Saliency Attribution (CHSA)} to dynamically prune the prefill context by quantifying each token’s contribution to latent reasoning, thereby retaining only the most informative elements for collaborative inference.

\noindent\textbf{Quantifying Reasoning Contribution.}
During ILD, attention matrices link latent states to original prefill tokens across layers and heads. To systematically evaluate a token's importance, we compute a global saliency score $S_j$ by first taking the maximum attention over all heads $H$ and layers $L$, and then averaging across latent reasoning steps $T$:
\begin{equation}
S_j = \frac{1}{T} \sum_{t=1}^{T} \left( \max_{l \in [1, L],   h \in [1, H]} \mathcal{A}_{t, l, h, j} \right)
\end{equation}
This procedure ensures that rare but critical cues—for instance, a distant pedestrian detected by a single head—are preserved, while the averaging over steps enforces temporal consistency and reduces the influence of transient activations. By combining information across multiple heads and layers, the method captures both localized and distributed features relevant to spatial reasoning and decision-making.

\noindent\textbf{KV Pruning and Construction.}
Once global saliency scores are computed, tokens are ranked by $S_j$ and the Top-$K$ most salient tokens are retained in their original sequential order. Their corresponding Key and Value vectors are extracted from the prefill cache to form the compressed salient cache $\mathcal{K}\mathcal{V}{salient}$. The final CHSA cache is then obtained by concatenating this salient context with the latent reasoning trace:
\begin{equation}
\mathcal{K}\mathcal{V}_{CHSA} = [\mathcal{K}\mathcal{V}_{salient}  \Vert\  \mathcal{K}\mathcal{V}_{latent}]
\end{equation}
By performing this selective pruning, CHSA distills the observation window to its most informative elements, significantly reducing V2V transmission load while preserving both spatial semantics and high-level decision-making intent. This mechanism enables ego-centric agents to communicate efficiently, maintaining critical environmental awareness and reasoning capabilities under strict bandwidth constraints.
\subsection{Shallow-Stream Knowledge Distillation (SSKD)}
\label{subsec:sskd}
While CHSA effectively reduces spatial redundancy, the fused reasoning trace still poses a significant risk to the ego vehicle’s decision stability. Inspired by Observation~\ref{ob2}, we propose \textbf{Shallow-Stream Knowledge Distillation (SSKD)}. We posit that the high-entropy Shallow-Stream provides pure spatial priors and intents that are universally beneficial, whereas the Deep-Stream contains the complex, task-specific synthesis that triggers Identity Confusion. SSKD acts as a structural filter, distilling the reasoning trace by truncating the complex Deep-Stream and retaining only the foundational Shallow-Stream. Specifically, we transmit only the first $L_{comm}$ layers. This effectively distills the collaborator’s internal states into a stable structured contextual prior, stripped of the perceptual artifacts and decision-making noise that reside in deeper layers:
\begin{equation}
    \mathcal{P} = \mathcal{K}\mathcal{V}_{CHSA}^{(1:L_{comm})}
\end{equation}
\subsection{Asymmetric Collaborative Inference}
Upon receiving $\mathcal{P}$ from a collaborator, the ego vehicle performs an asymmetric collaborative inference. For the shallow layers $l \leq L_{comm}$, the model incorporates collaborative context by concatenating local KV cache with the received stream:
\begin{equation}
\mathcal{KV}^{(l)}{fused} = [\mathcal{KV}{ego}^{(l)} \Vert\ \mathcal{P}^{(l)}]
\end{equation}
This allows the ego vehicle to integrate collaborator-derived perception and intent signals into its intermediate representation. For all subsequent deeper layers ($l > L_{comm}$), the model terminates external injection and relies exclusively on its own unadulterated internal states:
\begin{equation}
h^{(l)} = \text{TransformerBlock}^{(l)}(h^{(l-1)}, \mathcal{KV}_{ego}^{(l)})
\end{equation}
This asymmetric design ensures that the high-level action synthesis remains governed by the ego vehicle's independent state. The final action $a_t$ is conditioned on this enriched multi-agent context, achieving robust coordination without sacrificing decision autonomy or high efficiency.
\section{Experiment}
\label{sec:experiment}

\subsection{Experimental Setup}
\label{subsec:setup}
\begin{table*}[t]
\centering
\caption{Closed-Loop evaluation of different models across various communication methods. C/L refers to camera/LiDAR. NC: navigation command, TP: target point}
\label{tab:main_results}
\resizebox{\textwidth}{!}{
\begin{tabular}{l c c c l c c c c c c}
\toprule

\multirow{2}{*}{\textbf{Model}} & \multirow{2}{*}{\textbf{Cond.}} & \multirow{2}{*}{\textbf{Mod.}} & \textbf{Params} & \multirow{2}{*}{\textbf{Comm.}} & \multicolumn{2}{c}{\textbf{Vehicle 0 (V0)}} & \multicolumn{2}{c}{\textbf{Vehicle 1 (V1)}} & \multicolumn{2}{c}{\textbf{Communication}} \\
\cmidrule(lr){6-7} \cmidrule(lr){8-9} \cmidrule(lr){10-11}
& & & \textbf{(B)} & & \textbf{DS}~\textcolor{red}{$\uparrow$} & \textbf{RC}~\textcolor{red}{$\uparrow$} & \textbf{DS}~\textcolor{red}{$\uparrow$} & \textbf{RC}~\textcolor{red}{$\uparrow$} & \textbf{Latency}~(ms)~\textcolor{blue}{$\downarrow$} & \textbf{Size}~(KB)~\textcolor{blue}{$\downarrow$} \\
\midrule

\multirow{4}{*}{\textbf{ORION}} & \multirow{4}{*}{NC} & \multirow{4}{*}{C} & \multirow{4}{*}{7} 
& Noncollab & 26.68 & 52.12 & 25.91 & 50.32 & - & - \\
& & & & Language & 29.34 & 54.20 & 27.19 & 56.35 & 7802 & 2.3 \\
& & & & Visual & 31.48 & 60.48 & 29.88 & 59.82 & 82 & 8208 \\
& & & & \cellcolor{gray!15}\textbf{LACO} & \cellcolor{gray!15}\textbf{35.48} & \cellcolor{gray!15}\textbf{68.98} & \cellcolor{gray!15}\textbf{32.65} & \cellcolor{gray!15}\textbf{63.75} & \cellcolor{gray!15}430 & \cellcolor{gray!15}4881 \\
\midrule

\multirow{4}{*}{\textbf{SimLingo}} & \multirow{4}{*}{TP} & \multirow{4}{*}{C} & \multirow{4}{*}{0.5} 
& Noncollab & 28.58 & 56.70 & 23.35 & 51.63 & - & - \\
& & & & Language & 30.06 & 62.23 & 26.14 & 48.63 & 1300 & 1.8 \\
& & & & Visual & 32.14 & 65.03 & 27.24 & 52.17 & 52 & 896 \\
& & & & \cellcolor{gray!15}\textbf{LACO} & \cellcolor{gray!15}\textbf{35.73} & \cellcolor{gray!15}\textbf{72.06} & \cellcolor{gray!15}\textbf{29.22} & \cellcolor{gray!15}\textbf{68.00} & \cellcolor{gray!15}382 & \cellcolor{gray!15}103 \\
\midrule

\multirow{4}{*}{\textbf{LMDrive (LLaMA)}} & \multirow{4}{*}{NC} & \multirow{4}{*}{C\&L} & \multirow{4}{*}{7} 
& Noncollab & 15.33 & 23.29 & 19.82 & 35.30 & - & - \\
& & & & Language & 18.80 & 28.83 & 24.41 & 49.55 & 8509 & 2.51 \\
& & & & Visual & 19.32 & 37.20 & 29.09 & \textbf{65.15} & 95 & 334 \\
& & & & \cellcolor{gray!15}\textbf{LACO} & \cellcolor{gray!15}\textbf{22.84} & \cellcolor{gray!15}\textbf{39.55} & \cellcolor{gray!15}\textbf{30.54} & \cellcolor{gray!15}61.34 & \cellcolor{gray!15}215 & \cellcolor{gray!15}179 \\
\midrule

\multirow{4}{*}{\textbf{LMDrive (LLaVA)}} & \multirow{4}{*}{NC} & \multirow{4}{*}{C\&L} & \multirow{4}{*}{7} 
& Noncollab & 20.75 & 24.23 & 23.47 & 28.10 & - & - \\
& & & & Language & 24.82 & 28.51 & 24.46 & 35.54 & 8021 & 2.51 \\
& & & & Visual & 26.64 & 35.57 & 27.61 & 36.53 & 95 & 334 \\
& & & & \cellcolor{gray!15}\textbf{LACO} & \cellcolor{gray!15}\textbf{28.40} & \cellcolor{gray!15}\textbf{48.93} & \cellcolor{gray!15}\textbf{30.13} & \cellcolor{gray!15}\textbf{52.96} & \cellcolor{gray!15}215 & \cellcolor{gray!15}179 \\
\midrule

\multirow{4}{*}{\textbf{LMDrive (Vicuna)}} & \multirow{4}{*}{NC} & \multirow{4}{*}{C\&L} & \multirow{4}{*}{7} 
& Noncollab & 25.16 & 34.97 & 17.30 & 38.13 & - & - \\
& & & & Language & 24.79 & 38.03 & 24.25 & 32.20 & 8340 & 2.5 \\
& & & & Visual & 27.50 & 52.32 & 26.52 & 40.85 & 95 & 334 \\
& & & & \cellcolor{gray!15}\textbf{LACO} & \cellcolor{gray!15}\textbf{32.07} & \cellcolor{gray!15}\textbf{61.36} & \cellcolor{gray!15}\textbf{31.41} & \cellcolor{gray!15}\textbf{51.56} & \cellcolor{gray!15}203 & \cellcolor{gray!15}179 \\
\bottomrule

\end{tabular}
}
\end{table*}
\textbf{Simulation Environment.}
We conduct closed-loop evaluations in the CARLA simulator built upon LangCoop. Each scenario contains two Connected and Automated Vehicles (CAVs) governed by our framework, navigating complex urban environments populated with dynamic actors (vehicles, pedestrians, and cyclists) controlled by CARLA’s \cite{dosovitskiy2017carla} traffic manager. The two CAVs are initialized at different locations within the same vicinity to ensure meaningful interaction and potential collaboration. We assume a V2V communication range of 200 meters. For perception, each vehicle is equipped with a front-facing RGB camera capturing images at a resolution of $800 \times 600$.

\noindent\textbf{Evaluation Metrics.}
We adopt four metrics to comprehensively evaluate safety, efficiency, and communication overhead. Driving Score (DS) serves as the primary indicator of driving quality, defined as $DS = RC \times (1 - IP)$, where Route Completion (RC) measures the percentage of the predefined route successfully traversed (0–100\%), and Infraction Penalty (IP) aggregates penalties from collisions, traffic light violations, and lane invasions with severity-aware weighting. Communication Size quantifies the average data volume transmitted per interaction, while Communication Latency measures the end-to-end time required to generate and transmit the reasoning trace.

\noindent\textbf{Implementation Details and Baselines.}
We integrate our framework with ORION \cite{fu2025orion}, SimLingo \cite{renz2025simlingo}, and LMDrive \cite{shao2024lmdrive} (with various backbone including LLaMA \cite{touvron2023llama}, LLaVA \cite{liu2023visual} and Vicuna \cite{vicuna2023}) as representative VLA backbones, preserving their original architectures and pretrained weights to ensure fair comparison and retain their full representational and reasoning capacity. We compare against four collaboration paradigms: (1) Ego-only (Single-Agent), where each vehicle performs independent inference based solely on local observations without communication; (2) Language-based Collaboration \cite{gao2025langcoop,chiu2025v2v}, where agents exchange reasoning outputs and driving intents in natural language; (3) Visual-based Collaboration \cite{you2026v2x,gao2025stamp}, where agents directly share processed visual tokens to enlarge their perceptual receptive fields; and (4) LACO (ours), in which vehicles exchange distilled KV caches representing their internal thinking states.

For our latent-based approach, the number of iterative deliberation steps is set to $m=10$ to ensure sufficient depth for intent crystallization. For information pruning and distillation, we employ a CHSA retention ratio of $\rho=0.3$ and an SSKD distillation depth of $L_{comm}=10\%$. All of our experiments are conducted on 4 $\times$ RTX3090 GPUs.
\begin{figure*}[t]
    \centering
    \includegraphics[width=\linewidth]{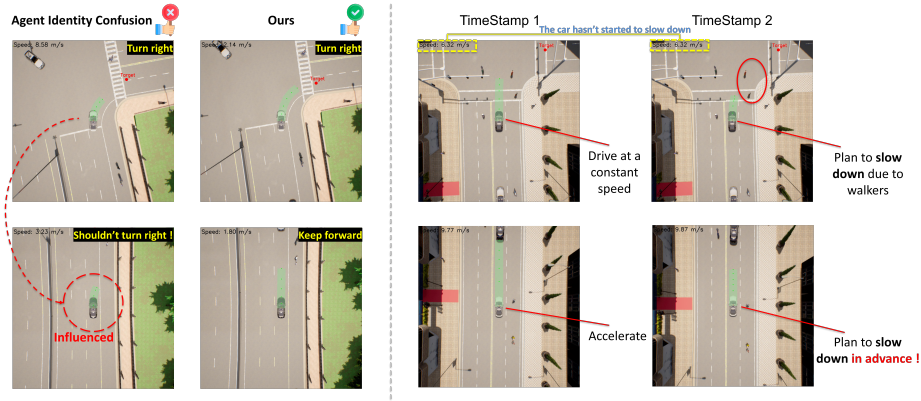}
    \caption{\textbf{Qualitative Visualization in closed-loop settings.} \textbf{(Left)} For naive KV fusion method, the agent suffers from identity confusion. LACO resolves this by selective knowledge distilling.
\textbf{(Right)} LACO enables the ego vehicle to anticipate the intentions and field of view of preceding vehicles, allowing it to proactively adjust speed and take precautionary actions before potential traffic conflicts arise.}
    \label{fig:qualitative}
\end{figure*}
\subsection{Quantitative Results}
\label{subsec:main_results}
\cref{tab:main_results} summarizes the closed-loop evaluation of our LACO framework against single-agent and collaborative baselines across multiple VLA architectures. The results demonstrate that LACO achieves state-of-the-art driving proficiency while strictly adhering to low latency and bandwidth constraints.

\noindent\textbf{Superior Driving Performance.} LACO consistently outperforms all baselines in Driving Score (DS) and Route Completion (RC). For instance, with the ORION backbone, LACO improves DS by an absolute margin of 8.80 over the Non-collaborative baseline and 4.00 over the Visual-based method. While Language-based collaboration suffers from lossy semantic compression and Visual-based sharing lacks explicit intent alignment, LACO's distilled latent trace encapsulates both spatial geometric priors and high-level decision rationale, significantly reducing collision rates and enhancing coordinated multi-agent progression.

\noindent\textbf{Latency and Bandwidth Efficiency.} A critical failure point of existing paradigms is their communication overhead. Language-based methods incur prohibitive autoregressive latency (\eg, $>8000$ ms for 7B models), rendering them infeasible for real-time driving. Conversely, Visual-based sharing circumvents latency but transmits uncompressed, high-dimensional tokens, resulting in severe bandwidth penalties (up to $8.2$ MB per exchange). Through non-autoregressive latent deliberation and layer-wise distillation, LACO bypasses the linguistic decoding bottleneck and explicitly prunes redundant features. Consequently, LACO accelerates end-to-end latency by $\sim 20\times$ compared to Language methods and slashes communication payloads by $40\% \sim 90\%$ compared to Visual sharing, all while delivering robust downstream performance across varying model scales (from 0.5B to 7B).
\begin{table*}[t]
    \centering
    \begin{minipage}[t]{0.45\textwidth}
        \centering
        \captionof{table}{Ablation study. ($^*$) Disabling ILD module inherently disables CHSA.}
        \label{tab:ablation}
        \renewcommand{\arraystretch}{1.75} 
        \resizebox{\linewidth}{!}{
        \begin{tabular}{ccc ccccccc}
        \toprule
        \multicolumn{3}{c}{\textbf{Modules}} & \multicolumn{2}{c}{\textbf{ORION}} & \multicolumn{2}{c}{\textbf{SimLingo}} & \multicolumn{2}{c}{\textbf{LMDrive}} \\
        \cmidrule(lr){1-3} \cmidrule(lr){4-5} \cmidrule(lr){6-7} \cmidrule(lr){8-9}
        \textbf{ILD} & \textbf{CHSA} & \textbf{SSKD} & \textbf{V0} & \textbf{V1} & \textbf{V0} & \textbf{V1} & \textbf{V0} & \textbf{V1} \\
        \midrule
        $\times$ & $\times$ & $\times$ & 30.05 & 27.14 & 31.48 & 24.34 & 23.77 & 25.15 \\
        $\times$ & $\times^*$ & $\checkmark$ & 33.95 & 31.34 & 34.95 & 28.16 & 27.85 & 28.88 \\
        $\checkmark$ & $\checkmark$ & $\times$ & 32.46 & 27.61 & 32.21 & 26.99 & 26.41 & 27.26 \\
        $\checkmark$ & $\times$(Full) & $\checkmark$ & \textbf{36.03} & 31.60 & 35.55 & \textbf{30.06} & 28.26 & 30.09 \\
        \rowcolor{gray!15} $\checkmark$ & $\checkmark$ & $\checkmark$ & 35.48 & \textbf{31.65} & \textbf{35.73} & 29.22 & \textbf{28.40} & \textbf{30.13} \\
        \bottomrule
        \end{tabular}
        }
    \end{minipage}
    \hfill
    \begin{minipage}[t]{0.54\textwidth}
        \centering

        \captionof{table}{SSKD distillation depth Analysis.}
        \label{tab:sskd_depth}
        \renewcommand{\arraystretch}{1.1} 
        \resizebox{\linewidth}{!}{
        \begin{tabular}{l cc cc cc cc cc}
        \toprule
        \multirow{2}{*}{\textbf{Model}} & \multicolumn{2}{c}{\textbf{5\%}} & \multicolumn{2}{c}{\cellcolor{gray!15}\textbf{10\% (Default)}} & \multicolumn{2}{c}{\cellcolor{gray!15}\textbf{30\%}} & \multicolumn{2}{c}{\textbf{50\%}} & \multicolumn{2}{c}{\textbf{Full}} \\
        \cmidrule(lr){2-3} \cmidrule(lr){4-5} \cmidrule(lr){6-7} \cmidrule(lr){8-9} \cmidrule(lr){10-11}
        & \textbf{V0} & \textbf{V1} & \cellcolor{gray!15}\textbf{V0} & \cellcolor{gray!15}\textbf{V1} & \cellcolor{gray!15}\textbf{V0} & \cellcolor{gray!15}\textbf{V1} & \textbf{V0} & \textbf{V1} & \textbf{V0} & \textbf{V1} \\
        \midrule
        \textbf{ORION}    & 34.04 & 32.14 & \cellcolor{gray!15}\textbf{35.48} & \cellcolor{gray!15}\textbf{32.65} & \cellcolor{gray!15}35.70 & \cellcolor{gray!15}32.60 & 34.70 & 30.43 & 32.46 & 27.61 \\
        \textbf{SimLingo} & 33.95 & 28.82 & \cellcolor{gray!15}35.73 & \cellcolor{gray!15}29.22 & \cellcolor{gray!15}\textbf{35.97} & \cellcolor{gray!15}\textbf{30.03} & 34.28 & 28.53 & 32.21 & 26.99 \\
        \textbf{LMDrive}  & 27.01 & 27.89 & \cellcolor{gray!15}\textbf{28.40} & \cellcolor{gray!15}30.13 & \cellcolor{gray!15}27.66 & \cellcolor{gray!15}\textbf{30.85} & 27.98 & 28.65 & 26.41 & 27.26 \\
        \bottomrule
        \end{tabular}
        }
        \captionof{table}{CHSA Retention Rate Analysis}
        \label{tab:chsa_ratio}
        \renewcommand{\arraystretch}{1.1}
        \resizebox{\linewidth}{!}{
        \begin{tabular}{l cc cc cc cc cc}
        \toprule
        \multirow{2}{*}{\textbf{Model}} & \multicolumn{2}{c}{\textbf{10\%}} & \multicolumn{2}{c}{\textbf{20\%}} & \multicolumn{2}{c}{\cellcolor{gray!15}\textbf{30\% (Default)}} & \multicolumn{2}{c}{\cellcolor{gray!15}\textbf{50\%}} & \multicolumn{2}{c}{\textbf{100\%}} \\
        \cmidrule(lr){2-3} \cmidrule(lr){4-5} \cmidrule(lr){6-7} \cmidrule(lr){8-9} \cmidrule(lr){10-11}
        & \textbf{V0} & \textbf{V1} & \textbf{V0} & \textbf{V1} & \cellcolor{gray!15}\textbf{V0} & \cellcolor{gray!15}\textbf{V1} & \cellcolor{gray!15}\textbf{V0} & \cellcolor{gray!15}\textbf{V1} & \textbf{V0} & \textbf{V1} \\
        \midrule
        \textbf{ORION}    & 31.05 & 28.58 & 33.53 & 30.08 & \cellcolor{gray!15}35.48 & \cellcolor{gray!15}31.65 & \cellcolor{gray!15}35.63 & \cellcolor{gray!15}\textbf{32.12} & \textbf{36.03} & 31.60 \\
        \textbf{SimLingo} & 33.98 & 26.50 & 35.71 & 27.93 & \cellcolor{gray!15}\textbf{35.73} & \cellcolor{gray!15}29.22 & \cellcolor{gray!15}35.49 & \cellcolor{gray!15}29.99 & 35.55 & \textbf{30.06} \\
        \textbf{LMDrive}  & 25.82 & 27.98 & 26.06 & 29.55 & \cellcolor{gray!15}28.40 & \cellcolor{gray!15}\textbf{30.13} & \cellcolor{gray!15}\textbf{28.83} & \cellcolor{gray!15}30.12 & 28.26 & 30.09 \\
        \bottomrule
        \end{tabular}
        }
    \end{minipage}
\end{table*}
\subsection{Qualitative Analysis}
\noindent\textbf{Clear Agent Identity Clarification.} As shown in the left side of \cref{fig:qualitative}, naive latent sharing \cite{zou2025latent,ye2025kvcomm} may entangle the ego vehicle's decision state with collaborator-specific intents, leading to unstable behaviors. 
With SSKD, LACO restricts communication to shallow spatial priors, preventing deep-layer task interference and preserving clear agent-specific decision boundaries.

\noindent\textbf{Proactive Planning via Global Perception.}
As shown in the right side of \cref{fig:qualitative}, LACO enables proactive coordination by exposing the ego vehicle to hazards beyond its field of view. 
Through ILD-based latent communication, the ego vehicle anticipates both occluded risks and upstream behavioral cues, enabling earlier and smoother planning responses.
\subsection{Ablation Study}
In this section, we conduct in-depth study of each component in LACO. Unless otherwise specified, we use LLaVA as the backbone for LMDrive. More experimental results are provided in the supplementary material.

\noindent\textbf{Effectiveness of Core Modules.} \cref{tab:ablation} validates the contribution of each component in LACO across three VLA backbones. Disabling SSKD yields the sharpest performance decrease, underscoring its critical role in preventing Agent Identity Confusion. Without SSKD, the ego vehicle's policy is easily hijacked by the collaborator's deep-layer task intents. Removing ILD also causes a notable performance drop, indicating that sharing crystallized reasoning traces—rather than raw visual features—is vital for proactive coordination. Although transmitting the uncompressed KV cache in some case yield higher scores, it represents an impractical upper bound that scales poorly with visual token resolution. CHSA achieves comparable performance while eliminating up to 70\% of spatial redundancy, demonstrating that collaborative driving decisions depend on only a sparse subset of reasoning-critical semantic anchors.

\begin{figure}[t]
    \centering
    \begin{minipage}[b]{0.48\textwidth}
        \centering
        \includegraphics[height=4cm, keepaspectratio]{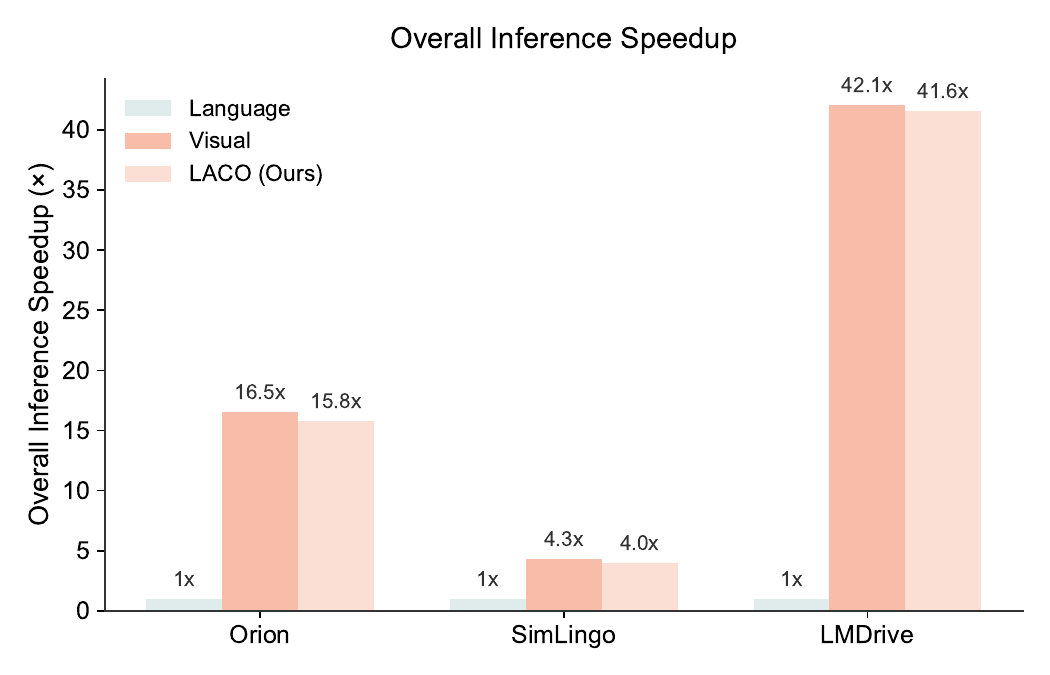}
        \caption{\textbf{Inference Speed-up Analysis.} LACO bypasses the autoregressive linguistic bottleneck by reusing prefilled latent states.}
        \label{fig:speedup}
    \end{minipage}
    \hfill
    \begin{minipage}[b]{0.48\textwidth}
        \centering
        \includegraphics[height=4cm,keepaspectratio]{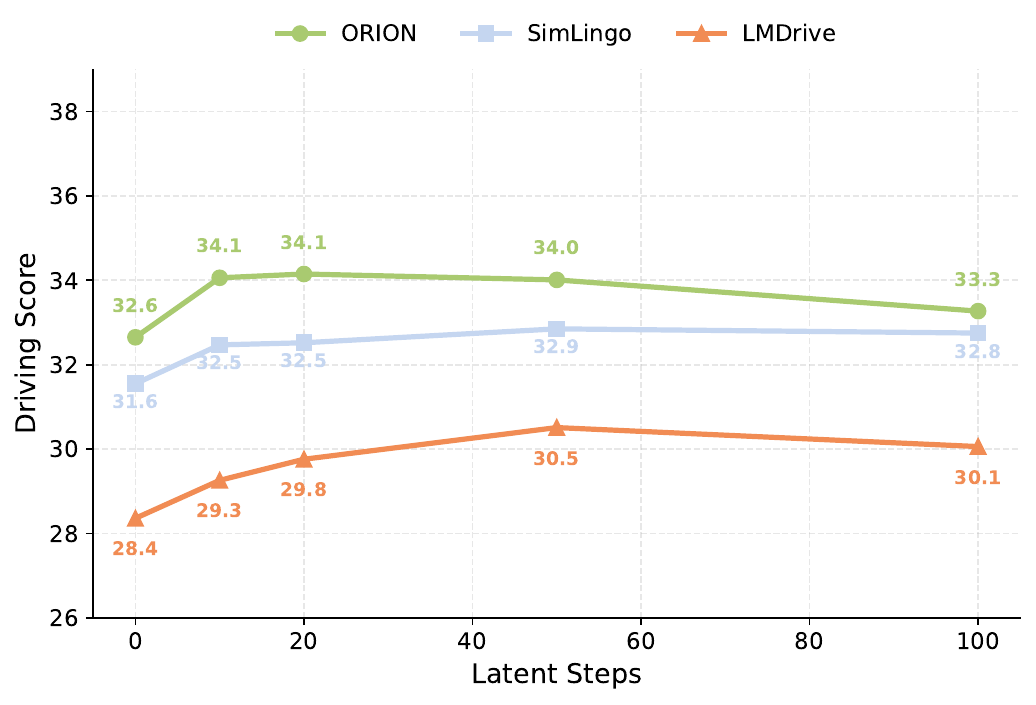}
        \caption{\textbf{Impact of the reasoning horizon.} A moderate deliberation phase optimizes the crystallization of raw perception into collaborative intent.}
        \label{fig:latentstep}
    \end{minipage}
\end{figure}

\noindent\textbf{Latent Prefilling Eliminates Additional Collaborative Overhead.} \cref{fig:speedup} evaluates the overall inference speed-up of LACO relative to the pure Language and standard Visual baselines. While the text-only configuration suffers from severe latency bottlenecks caused by autorgressive process, directly transferring visual tokens significantly accelerates the inference process. Crucially, despite introducing complex multi-agent communication, LACO incurs only a marginal speed reduction compared to the naive single-agent Visual baseline. This highly optimized efficiency is achieved by leveraging the initial latent reasoning stage as a prefill mechanism. Rather than repeatedly processing cumbersome raw visual tokens during collaboration, LACO caches the distilled latent representations. Subsequent spatial pruning and inter-agent communication operate strictly on these prefilled hidden states, completely circumventing the overhead of redundant visual information processing.

\noindent\textbf{Moderate Latent Deliberation Optimizes Intent Crystallization.} \cref{fig:latentstep} investigate the impact of the reasoning horizon by varying the number of iterative latent steps within the ILD module. Transitioning from a baseline of zero latent steps,which inherently restricts the system to naive perception sharing, to a moderate reasoning horizon yields a substantial performance leap across all VLA backbones. This confirms that a proper period of internal deliberation is crucial for crystallizing raw visual features into actionable, high-level driving intents. However, excessively prolonging the deliberation phase offers diminishing returns and sometimes leads to performance degradation. We attribute this decline to semantic drift or over-reasoning \cite{su2025between}: excessive autoregressive latent generation without continuous visual grounding can accumulate hallucinated noise, eventually polluting the shared collaborative intent.

\noindent\textbf{Optimal Distillation Depth Bridges Perception and Intent.} \cref{tab:sskd_depth} examines the impact of progressively increasing the shared KV depth in SSKD. Extremely shallow sharing yields limited improvement, as early-layer representations are still dominated by low-level perceptual encoding and lack higher-level situational structure. 
Performance improves substantially when the shared depth falls within the 10\%–30\% range, where different VLA backbones achieve their respective optima. This intermediate regime corresponds to a stage in which latent states encode globally grounded situational reasoning together with emerging intent representations, while remaining structurally disentangled from finalized ego-centric control commitments. Such representations enable effective cross-agent awareness injection without perturbing the receiver’s decision head.
However, further extending the shared depth leads to consistent performance degradation. At late stages, representations become tightly coupled with ego-specific action synthesis. Injecting these entangled control-oriented states introduces decision-level interference, manifesting as \textit{Agent Identity Confusion}.

\noindent\textbf{Reasonable Compression Reduces Bandwidth While Maintaining Performance.}
We examine the influence of different retention rates in CHSA, as shown in \cref{tab:chsa_ratio}. Across multiple VLA backbones, increasing the retention rate from 10\% to 30--50\% consistently yields substantial performance improvements (performance gain $\simeq 2$), indicating that the majority of decision-critical visual information is concentrated within a relatively small subset of high-saliency tokens. However, further increasing the retention rate beyond this range brings only marginal gains ($\simeq 0.5$) and the improvements become inconsistent across models, revealing a clear diminishing-return pattern. This observation suggests that most additional low-saliency tokens contribute marginally to downstream decision-making once the key semantic anchors are preserved. CHSA therefore effectively isolates the structurally dominant visual tokens required for control synthesis. 
\section{Conclusion}
We presented LACO, a latent-level collaboration framework for multi-agent Vision-Language-Action driving models. 
By identifying Agent Identity Confusion as a key limitation of naive deep latent sharing, we introduced Shallow-Stream Knowledge Distillation to prevent deep decision-state entanglement while enabling structured sharing of spatial and intent representations. Combined with Iterative Latent Deliberation and Cross-Horizon Saliency Attribution, LACO enables latency-aware, bandwidth-efficient latent communication without redundant token processing. 
Closed-loop evaluations demonstrate that effective multi-agent coordination can emerge from selective internal representation sharing and achieve remarkable performance.

%
%
\bibliographystyle{splncs04}
\bibliography{main}

\begin{thebibliography}{10}
\providecommand{\url}[1]{\texttt{#1}}
\providecommand{\urlprefix}{URL }
\providecommand{\doi}[1]{https://doi.org/#1}

\bibitem{arnold2020cooperative}
Arnold, E., Dianati, M., De~Temple, R., Fallah, S.: Cooperative perception for 3d object detection in driving scenarios using infrastructure sensors. IEEE Transactions on Intelligent Transportation Systems  \textbf{23}(3),  1852--1864 (2020)

\bibitem{chen2019cooper}
Chen, Q., Tang, S., Yang, Q., Fu, S.: Cooper: Cooperative perception for connected autonomous vehicles based on 3d point clouds. In: 2019 IEEE 39th International Conference on distributed computing systems (ICDCS). pp. 514--524. IEEE (2019)

\bibitem{vicuna2023}
Chiang, W.L., Li, Z., Lin, Z., Sheng, Y., Wu, Z., Zhang, H., Zheng, L., Zhuang, S., Zhuang, Y., Gonzalez, J.E., Stoica, I., Xing, E.P.: Vicuna: An open-source chatbot impressing gpt-4 with 90\%* chatgpt quality (March 2023), \url{https://lmsys.org/blog/2023-03-30-vicuna/}

\bibitem{chiu2025v2v}
Chiu, H.k., Hachiuma, R., Wang, C.Y., Smith, S.F., Wang, Y.C.F., Chen, M.H.: V2v-llm: Vehicle-to-vehicle cooperative autonomous driving with multi-modal large language models. arXiv preprint arXiv:2502.09980  (2025)

\bibitem{dery2026latent}
Dery, L.M., Yahav, Z., Prior, H., Feng, Q., Shen, J., Szlam, A.: Latent space communication via kv cache alignment. arXiv preprint arXiv:2601.06123  (2026)

\bibitem{dosovitskiy2017carla}
Dosovitskiy, A., Ros, G., Codevilla, F., Lopez, A., Koltun, V.: Carla: An open urban driving simulator. In: Conference on robot learning. pp. 1--16. PMLR (2017)

\bibitem{du2025enabling}
Du, Z., Wang, R., Bai, H., Cao, Z., Zhu, X., Cheng, Y., Zheng, B., Chen, W., Ying, H.: Enabling agents to communicate entirely in latent space. arXiv preprint arXiv:2511.09149  (2025)

\bibitem{fu2025orion}
Fu, H., Zhang, D., Zhao, Z., Cui, J., Liang, D., Zhang, C., Zhang, D., Xie, H., Wang, B., Bai, X.: Orion: A holistic end-to-end autonomous driving framework by vision-language instructed action generation. In: Proceedings of the IEEE/CVF International Conference on Computer Vision. pp. 24823--24834 (2025)

\bibitem{fu2025cache}
Fu, T., Min, Z., Zhang, H., Yan, J., Dai, G., Ouyang, W., Wang, Y.: Cache-to-cache: Direct semantic communication between large language models. arXiv preprint arXiv:2510.03215  (2025)

\bibitem{gao2025langcoop}
Gao, X., Wu, Y., Wang, R., Liu, C., Zhou, Y., Tu, Z.: Langcoop: Collaborative driving with language. In: Proceedings of the Computer Vision and Pattern Recognition Conference. pp. 4226--4237 (2025)

\bibitem{gao2025stamp}
Gao, X., Xu, R., Li, J., Wang, Z., Fan, Z., Tu, Z.: Stamp: Scalable task and model-agnostic collaborative perception. arXiv preprint arXiv:2501.18616  (2025)

\bibitem{guo2024large}
Guo, T., Chen, X., Wang, Y., Chang, R., Pei, S., Chawla, N.V., Wiest, O., Zhang, X.: Large language model based multi-agents: A survey of progress and challenges. arXiv preprint arXiv:2402.01680  (2024)

\bibitem{han2023collaborative}
Han, Y., Zhang, H., Li, H., Jin, Y., Lang, C., Li, Y.: Collaborative perception in autonomous driving: Methods, datasets, and challenges. IEEE Intelligent Transportation Systems Magazine  \textbf{15}(6),  131--151 (2023)

\bibitem{hao2024training}
Hao, S., Sukhbaatar, S., Su, D., Li, X., Hu, Z., Weston, J., Tian, Y.: Training large language models to reason in a continuous latent space. arXiv preprint arXiv:2412.06769  (2024)

\bibitem{hong2023metagpt}
Hong, S., Zhuge, M., Chen, J., Zheng, X., Cheng, Y., Wang, J., Zhang, C., Wang, Z., Yau, S.K.S., Lin, Z., et~al.: Metagpt: Meta programming for a multi-agent collaborative framework. In: The twelfth international conference on learning representations (2023)

\bibitem{hu2022where2comm}
Hu, Y., Fang, S., Lei, Z., Zhong, Y., Chen, S.: Where2comm: Communication-efficient collaborative perception via spatial confidence maps. Advances in neural information processing systems  \textbf{35},  4874--4886 (2022)

\bibitem{hwang2024emma}
Hwang, J.J., Xu, R., Lin, H., Hung, W.C., Ji, J., Choi, K., Huang, D., He, T., Covington, P., Sapp, B., et~al.: Emma: End-to-end multimodal model for autonomous driving. arXiv preprint arXiv:2410.23262  (2024)

\bibitem{jiang2024koma}
Jiang, K., Cai, X., Cui, Z., Li, A., Ren, Y., Yu, H., Yang, H., Fu, D., Wen, L., Cai, P.: Koma: Knowledge-driven multi-agent framework for autonomous driving with large language models. IEEE Transactions on Intelligent Vehicles  (2024)

\bibitem{jiang2025survey}
Jiang, S., Huang, Z., Qian, K., Luo, Z., Zhu, T., Zhong, Y., Tang, Y., Kong, M., Wang, Y., Jiao, S., et~al.: A survey on vision-language-action models for autonomous driving. In: Proceedings of the IEEE/CVF International Conference on Computer Vision. pp. 4524--4536 (2025)

\bibitem{karypidis2024dino}
Karypidis, E., Kakogeorgiou, I., Gidaris, S., Komodakis, N.: Dino-foresight: Looking into the future with dino. arXiv preprint arXiv:2412.11673  (2024)

\bibitem{kwon2023efficient}
Kwon, W., Li, Z., Zhuang, S., Sheng, Y., Zheng, L., Yu, C.H., Gonzalez, J., Zhang, H., Stoica, I.: Efficient memory management for large language model serving with pagedattention. In: Proceedings of the 29th symposium on operating systems principles. pp. 611--626 (2023)

\bibitem{li2025recogdrive}
Li, Y., Xiong, K., Guo, X., Li, F., Yan, S., Xu, G., Zhou, L., Chen, L., Sun, H., Wang, B., et~al.: Recogdrive: A reinforced cognitive framework for end-to-end autonomous driving. arXiv preprint arXiv:2506.08052  (2025)

\bibitem{liu2023visual}
Liu, H., Li, C., Wu, Q., Lee, Y.J.: Visual instruction tuning. Advances in neural information processing systems  \textbf{36},  34892--34916 (2023)

\bibitem{liu2020when2com}
Liu, Y.C., Tian, J., Glaser, N., Kira, Z.: When2com: Multi-agent perception via communication graph grouping. In: Proceedings of the IEEE/CVF Conference on computer vision and pattern recognition. pp. 4106--4115 (2020)

\bibitem{renz2025simlingo}
Renz, K., Chen, L., Arani, E., Sinavski, O.: Simlingo: Vision-only closed-loop autonomous driving with language-action alignment. In: Proceedings of the Computer Vision and Pattern Recognition Conference. pp. 11993--12003 (2025)

\bibitem{saunshi2025reasoning}
Saunshi, N., Dikkala, N., Li, Z., Kumar, S., Reddi, S.J.: Reasoning with latent thoughts: On the power of looped transformers. arXiv preprint arXiv:2502.17416  (2025)

\bibitem{shao2024lmdrive}
Shao, H., Hu, Y., Wang, L., Song, G., Waslander, S.L., Liu, Y., Li, H.: Lmdrive: Closed-loop end-to-end driving with large language models. In: Proceedings of the IEEE/CVF conference on computer vision and pattern recognition. pp. 15120--15130 (2024)

\bibitem{sima2024drivelm}
Sima, C., Renz, K., Chitta, K., Chen, L., Zhang, H., Xie, C., Bei{\ss}wenger, J., Luo, P., Geiger, A., Li, H.: Drivelm: Driving with graph visual question answering. In: European conference on computer vision. pp. 256--274. Springer (2024)

\bibitem{su2025between}
Su, J., Healey, J., Nakov, P., Cardie, C.: Between underthinking and overthinking: An empirical study of reasoning length and correctness in llms. arXiv preprint arXiv:2505.00127  (2025)

\bibitem{tan2025think}
Tan, W., Li, J., Ju, J., Luo, Z., Song, R., Luan, J.: Think silently, think fast: Dynamic latent compression of llm reasoning chains. arXiv preprint arXiv:2505.16552  (2025)

\bibitem{touvron2023llama}
Touvron, H., Lavril, T., Izacard, G., Martinet, X., Lachaux, M.A., Lacroix, T., Rozi{\`e}re, B., Goyal, N., Hambro, E., Azhar, F., et~al.: Llama: Open and efficient foundation language models. arXiv preprint arXiv:2302.13971  (2023)

\bibitem{vaswani2017attention}
Vaswani, A., Shazeer, N., Parmar, N., Uszkoreit, J., Jones, L., Gomez, A.N., Kaiser, {\L}., Polosukhin, I.: Attention is all you need. Advances in neural information processing systems  \textbf{30} (2017)

\bibitem{wang2020v2vnet}
Wang, T.H., Manivasagam, S., Liang, M., Yang, B., Zeng, W., Urtasun, R.: V2vnet: Vehicle-to-vehicle communication for joint perception and prediction. In: European conference on computer vision. pp. 605--621. Springer (2020)

\bibitem{wang2025alpamayo}
Wang, Y., Luo, W., Bai, J., Cao, Y., Che, T., Chen, K., Chen, Y., Diamond, J., Ding, Y., Ding, W., et~al.: Alpamayo-r1: Bridging reasoning and action prediction for generalizable autonomous driving in the long tail. arXiv preprint arXiv:2511.00088  (2025)

\bibitem{wu2025v2x}
Wu, K., Li, P., Zhou, Y., Gan, R., You, J., Cheng, Y., Zhu, J., Parker, S.T., Ran, B., Noyce, D.A., et~al.: V2x-llm: Enhancing v2x integration and understanding in connected vehicle corridors. arXiv preprint arXiv:2503.02239  (2025)

\bibitem{wu2025background}
Wu, Y., Gao, X., Tau, Q., Tu, Z., Lee, D.: Background fades, foreground leads: Curriculum-guided background pruning for efficient foreground-centric collaborative perception. arXiv preprint arXiv:2510.19250  (2025)

\bibitem{xu2025cosdh}
Xu, J., Zhang, Y., Cai, Z., Huang, D.: Cosdh: communication-efficient collaborative perception via supply-demand awareness and intermediate-late hybridization. In: Proceedings of the Computer Vision and Pattern Recognition Conference. pp. 6834--6843 (2025)

\bibitem{xu2022v2x}
Xu, R., Xiang, H., Tu, Z., Xia, X., Yang, M.H., Ma, J.: V2x-vit: Vehicle-to-everything cooperative perception with vision transformer. In: European conference on computer vision. pp. 107--124. Springer (2022)

\bibitem{xu2022opv2v}
Xu, R., Xiang, H., Xia, X., Han, X., Li, J., Ma, J.: Opv2v: An open benchmark dataset and fusion pipeline for perception with vehicle-to-vehicle communication. In: 2022 International Conference on Robotics and Automation (ICRA). pp. 2583--2589. IEEE (2022)

\bibitem{yang2025drivemoe}
Yang, Z., Chai, Y., Jia, X., Li, Q., Shao, Y., Zhu, X., Su, H., Yan, J.: Drivemoe: Mixture-of-experts for vision-language-action model in end-to-end autonomous driving. arXiv preprint arXiv:2505.16278  (2025)

\bibitem{ye2025kvcomm}
Ye, H., Gao, Z., Ma, M., Wang, Q., Fu, Y., Chung, M.Y., Lin, Y., Liu, Z., Zhang, J., Zhuo, D., et~al.: Kvcomm: Online cross-context kv-cache communication for efficient llm-based multi-agent systems. arXiv preprint arXiv:2510.12872  (2025)

\bibitem{you2026v2x}
You, J., Jiang, Z., Huang, Z., Shi, H., Gan, R., Wu, K., Cheng, X., Li, X., Ran, B.: V2x-vlm: End-to-end v2x cooperative autonomous driving through large vision-language models. Transportation Research Part C: Emerging Technologies  \textbf{183},  105457 (2026)

\bibitem{zhang2024chain}
Zhang, Y., Sun, R., Chen, Y., Pfister, T., Zhang, R., Arik, S.: Chain of agents: Large language models collaborating on long-context tasks. Advances in Neural Information Processing Systems  \textbf{37},  132208--132237 (2024)

\bibitem{zhu2025scaling}
Zhu, R.J., Wang, Z., Hua, K., Zhang, T., Li, Z., Que, H., Wei, B., Wen, Z., Yin, F., Xing, H., et~al.: Scaling latent reasoning via looped language models. arXiv preprint arXiv:2510.25741  (2025)

\bibitem{zou2025latent}
Zou, J., Yang, X., Qiu, R., Li, G., Tieu, K., Lu, P., Shen, K., Tong, H., Choi, Y., He, J., et~al.: Latent collaboration in multi-agent systems. arXiv preprint arXiv:2511.20639  (2025)

\end{thebibliography}
\clearpage
\setcounter{page}{1}
\setcounter{linenumber}{1} 
\setcounter{section}{1} 
\setcounter{equation}{0}
\setcounter{table}{0}
\setcounter{figure}{0}
\appendix
\begin{center}
    \Large \textbf{LACO: Adaptive Latent Communication for Collaborative Driving} \\
    \large \textit{Supplementary Material}
\end{center}
\noindent
This supplementary material provides additional details to support the main paper. 
Specifically, it includes the following components:

\begin{itemize}
\item \textbf{Implementation Details}: including the benchmark environment, evaluation metrics, and detailed descriptions of method implementations.
\item \textbf{Further Experiments}: additional experimental results and analyses that complement the findings in the main paper.
\end{itemize}

\section{Implementation Detail}
\subsection{Evaluation Environment and Scenarios}

We conduct our experiments on the LangCoop benchmark built upon the CARLA simulator. 
The benchmark provides diverse urban driving environments and traffic configurations 
for evaluating collaborative autonomous driving systems.

\begin{figure}[t]
\centering
\includegraphics[width=0.5\linewidth]{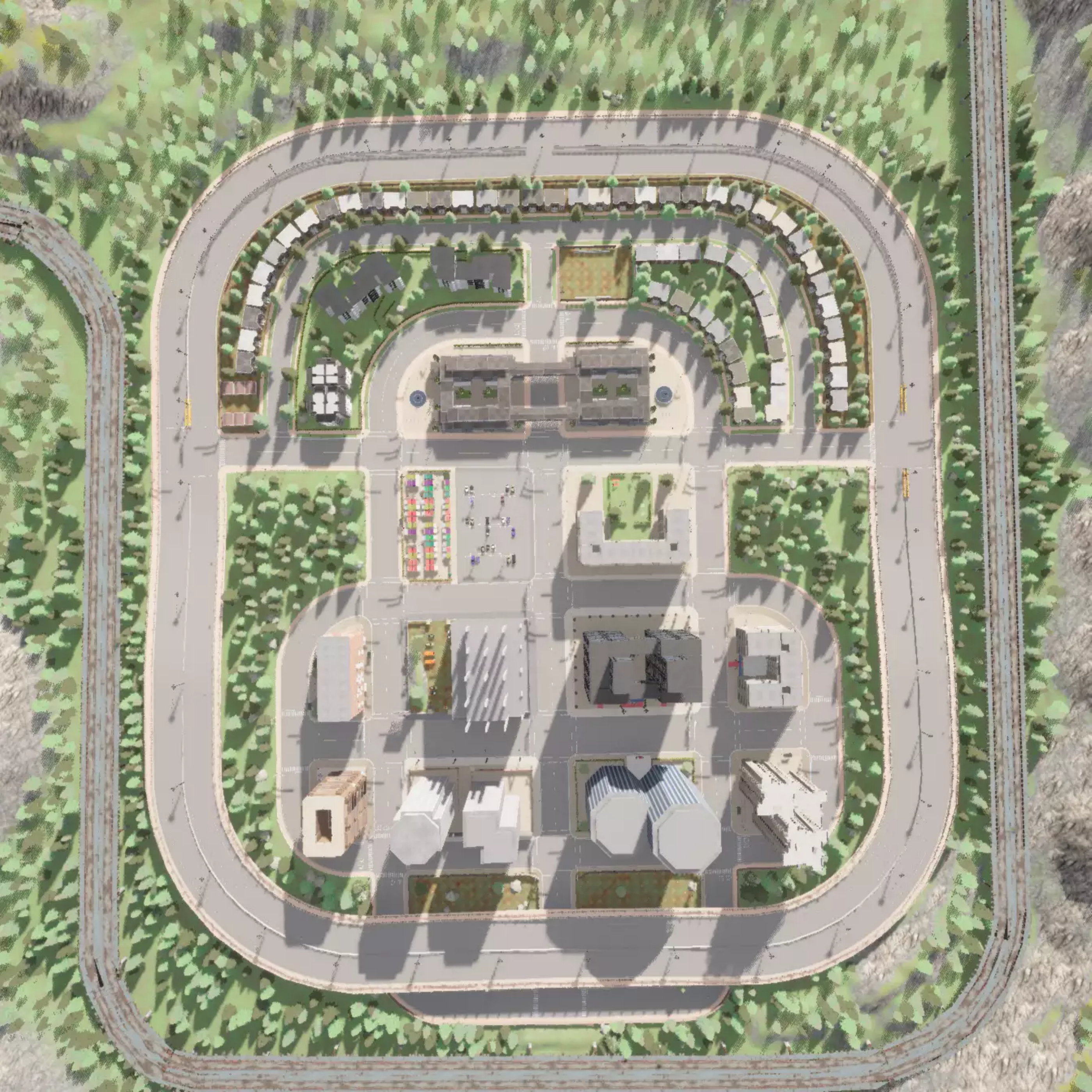}
\caption{Overview of the Town05 evaluation map used in the LangCoop benchmark. 
The map contains multiple predefined routes and diverse urban road structures 
including intersections, roundabouts, and multi-lane roads, providing 
challenging scenarios for evaluating collaborative autonomous driving systems.}
\label{fig:town05_map}
\end{figure}

Following the standard benchmark protocol, the final evaluation is conducted on 
Town05, as shown in \cref{fig:town05_map}, which contains a set of predefined test routes covering multiple 
traffic configurations. These configurations vary the number of vehicles and pedestrians, 
the probability of pedestrians violating traffic rules, and the trigger distance of 
event-based scenarios.

The benchmark also includes a variety of trigger-based scenarios, such as pedestrians 
or cyclists suddenly emerging from occluded areas and complex interactions at 
intersections. These scenarios create highly dynamic and partially observable 
traffic conditions, allowing us to evaluate the robustness and safety of the 
autonomous driving system.
\subsection{Metric Description}

We evaluate autonomous driving performance using the standard protocol adopted by the Langcoop benchmark. Each route is evaluated independently using three metrics: \textit{Driving Score}, \textit{Route Completion}, and \textit{Infraction Score}. After all routes are completed, the final metrics are obtained by averaging the per-route results.

\subsubsection{Driving Score. }

The primary evaluation metric is the \textit{Driving Score (DS)}, defined as

\begin{equation}
DS_i = RC_i \cdot IS_i ,
\end{equation}

where $RC_i$ denotes the \textit{Route Completion} percentage for route $i$, and $IS_i$ denotes the \textit{Infraction Score}, which penalizes traffic rule violations incurred during the route.

The driving score therefore reflects both task completion performance and driving safety compliance.

\subsubsection{Route Completion. }

The \textit{Route Completion (RC)} measures the fraction of the predefined route successfully traversed by the agent.

Let $d_{\text{traversed}}$ denote the cumulative distance traveled along the route and $d_{\text{total}}$ denote the total route length. The route completion score is defined as

\begin{equation}
RC_i =
\frac{d_{\text{traversed}}}{d_{\text{total}}} \times 100 .
\end{equation}

During evaluation, the vehicle's position is continuously compared with upcoming route waypoints. A waypoint is considered passed when the vehicle crosses the waypoint plane along the waypoint’s forward direction.

A route is considered successfully completed when the vehicle has already traversed more than \textbf{40\%} of the route and reaches within \textbf{10 meters} of the final target location.
\subsubsection{Infraction Penalty. }

Traffic violations reduce the driving score through the \textit{Infraction Score (IS)}. For route $i$, the infraction score is computed as the product of penalty coefficients for each type of infraction:

\begin{equation}
IS_i =
\prod_{j=1}^{N_I} (p_j)^{n_j},
\end{equation}

where $p_j$ denotes the penalty coefficient for the $j$-th infraction type, $n_j$ denotes the number of times that infraction occurred, and $N_I$ denotes the number of infraction categories.

The calculation starts with a base score of $1.0$, which is multiplicatively reduced whenever infractions occur.

\subsubsection{Infraction Types and Penalties. }

The evaluation considers several traffic infractions, each associated with a penalty coefficient:

\begin{table}[h]
\centering
\begin{tabular}{lc}
\hline
Infraction Type & Penalty Coefficient \\
\hline
Collision with pedestrians & 0.50 \\
Collision with vehicles & 0.60 \\
Collision with static objects & 0.65 \\
Running a red light & 0.70 \\
Ignoring a stop sign & 0.80 \\
Scenario timeout & 0.70 \\
Failure to yield to emergency vehicles & 0.70 \\
\hline
\end{tabular}
\caption{Infraction penalty coefficients used in the evaluation.}
\end{table}

Each occurrence multiplies the current infraction score by its corresponding penalty coefficient.

In addition, lane departure behavior is tracked and reported as the proportion of the traveled route distance spent outside valid route lanes.

\subsubsection{Route Termination Conditions. }

A route evaluation terminates immediately if one of the following events occurs:

\begin{itemize}
\item Route deviation (the vehicle deviates more than \textbf{50 meters} from the reference route).
\item Blocked agent (the vehicle remains below \textbf{0.5 m/s} for more than \textbf{30 seconds}).
\item Communication timeout (the agent fails to produce control commands within \textbf{60 seconds}).
\item Route timeout (the simulation exceeds the maximum allowed time for the route).
\end{itemize}

When a route terminates prematurely, the current route completion percentage is used to compute the final evaluation metrics. All infractions accumulated before termination are still applied when computing the final \textit{Driving Score}.

\subsubsection{Final Benchmark Score. }

After evaluating all routes, the final benchmark metrics are obtained by averaging the per-route scores:

\begin{equation}
DS = \frac{1}{N}\sum_{i=1}^{N} DS_i
\end{equation}

\begin{equation}
RC = \frac{1}{N}\sum_{i=1}^{N} RC_i
\end{equation}

where $N$ denotes the number of evaluated routes.

The \textit{Driving Score} is used as the primary ranking metric in the benchmark.

\subsection{Sensor Configuration}
\begin{figure}[t]
\centering

\begin{subfigure}{0.48\linewidth}
\centering
\includegraphics[width=\linewidth]{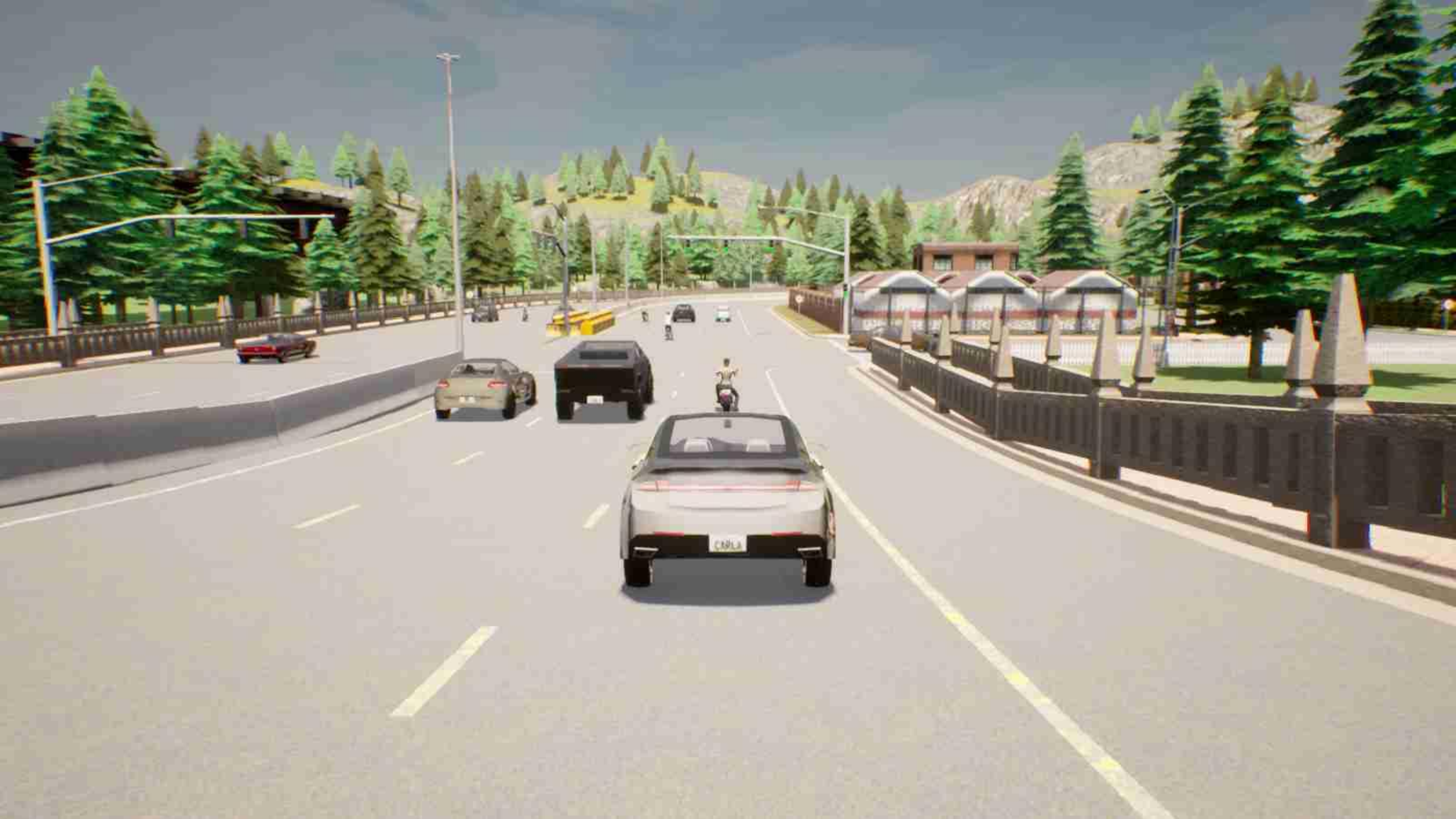}
\caption{Front}
\end{subfigure}
\hfill
\begin{subfigure}{0.48\linewidth}
\centering
\includegraphics[width=\linewidth]{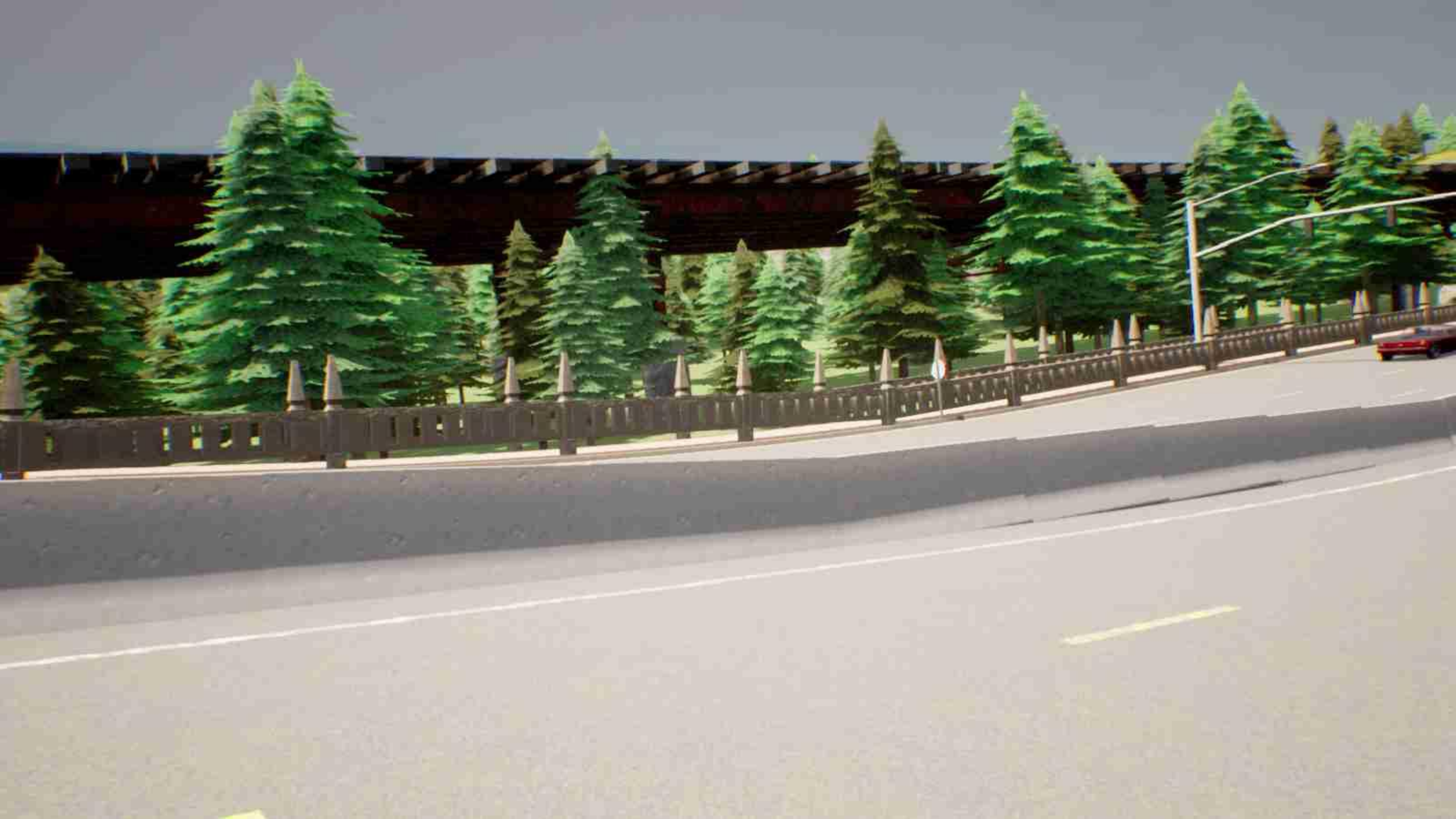}
\caption{Front Left}
\end{subfigure}

\begin{subfigure}{0.48\linewidth}
\centering
\includegraphics[width=\linewidth]{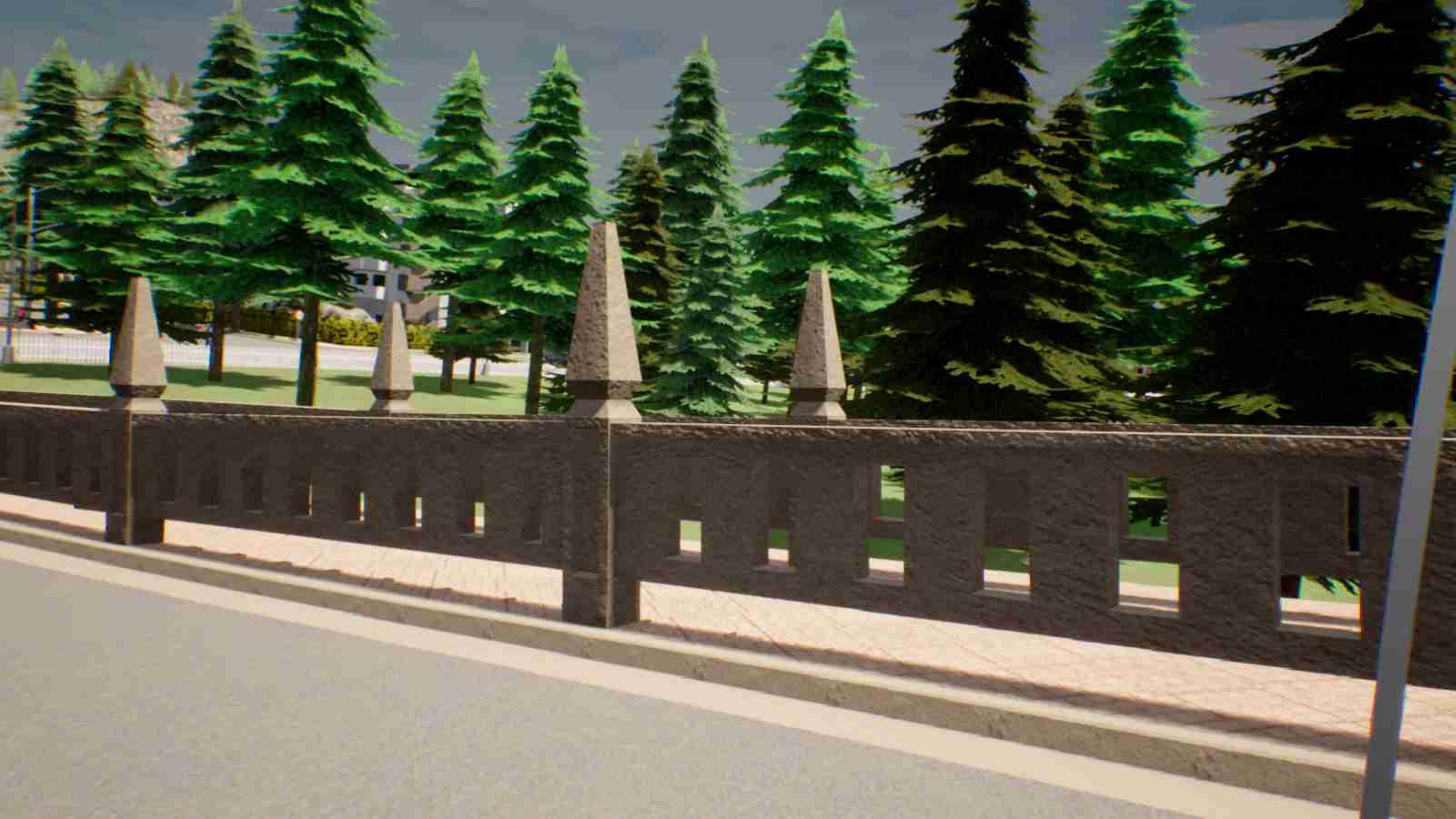}
\caption{Front Right}
\end{subfigure}
\hfill
\begin{subfigure}{0.48\linewidth}
\centering
\includegraphics[width=\linewidth]{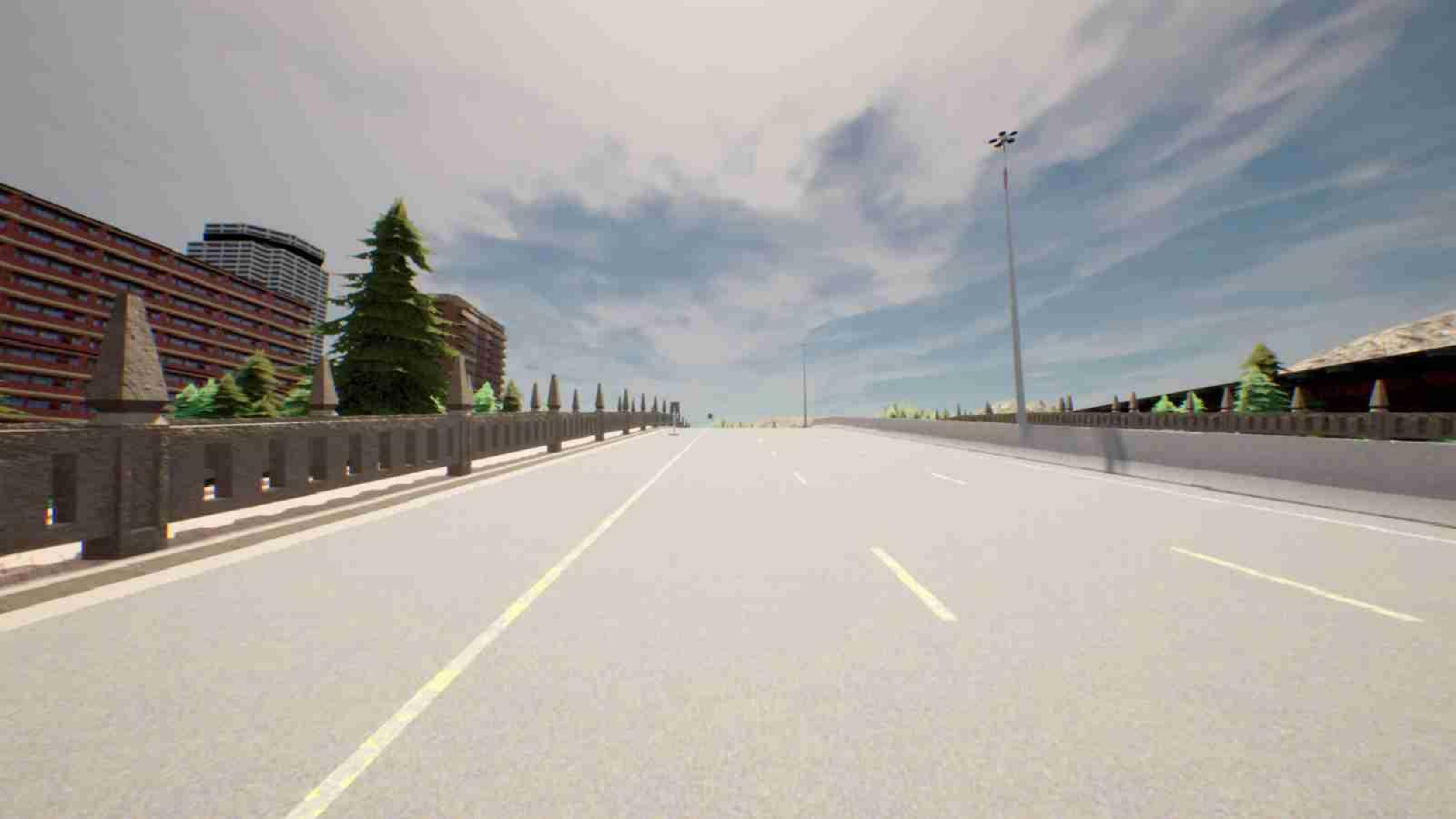}
\caption{Rear}
\end{subfigure}

\begin{subfigure}{0.48\linewidth}
\centering
\includegraphics[width=\linewidth]{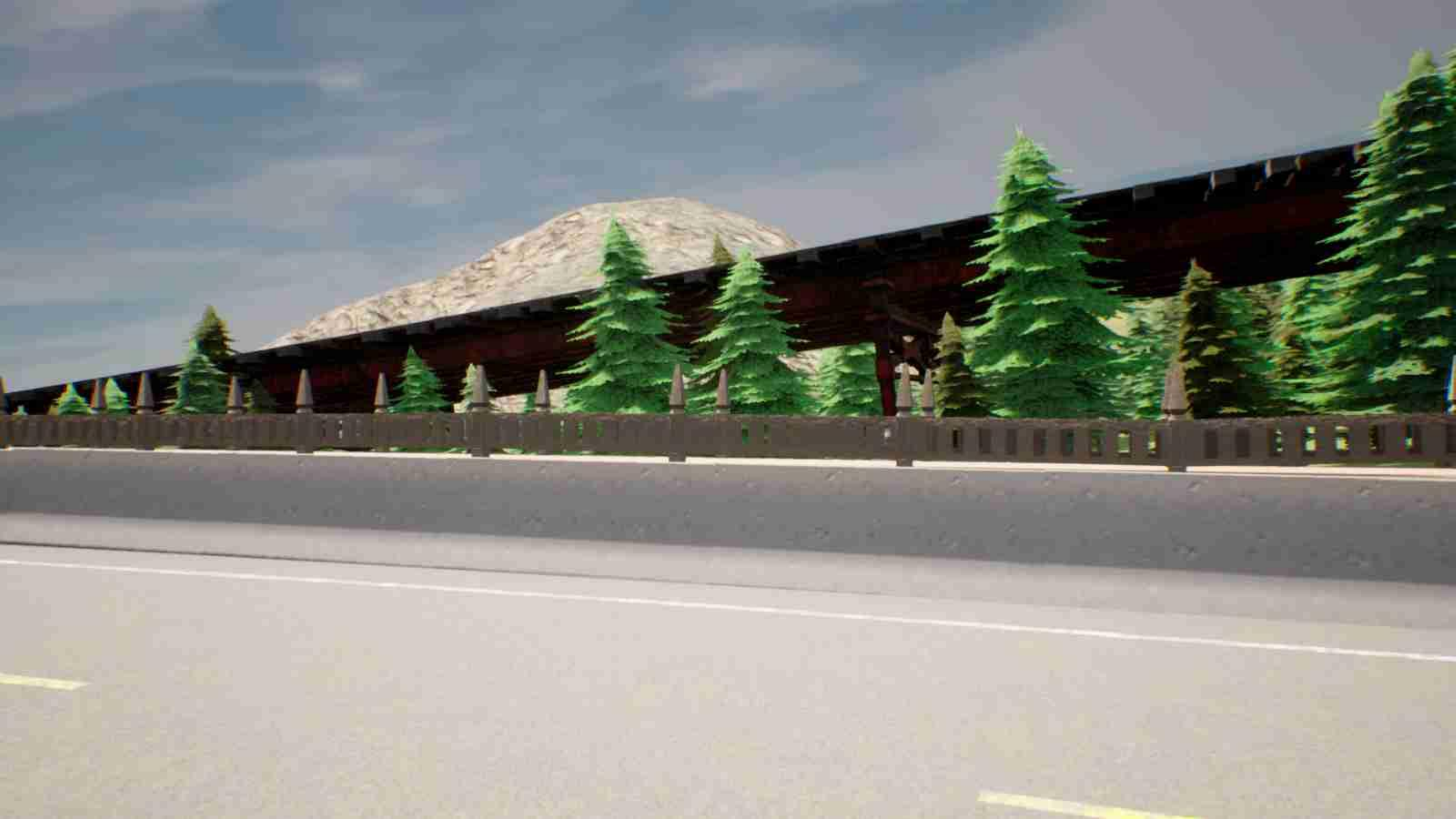}
\caption{Rear Left}
\end{subfigure}
\hfill
\begin{subfigure}{0.48\linewidth}
\centering
\includegraphics[width=\linewidth]{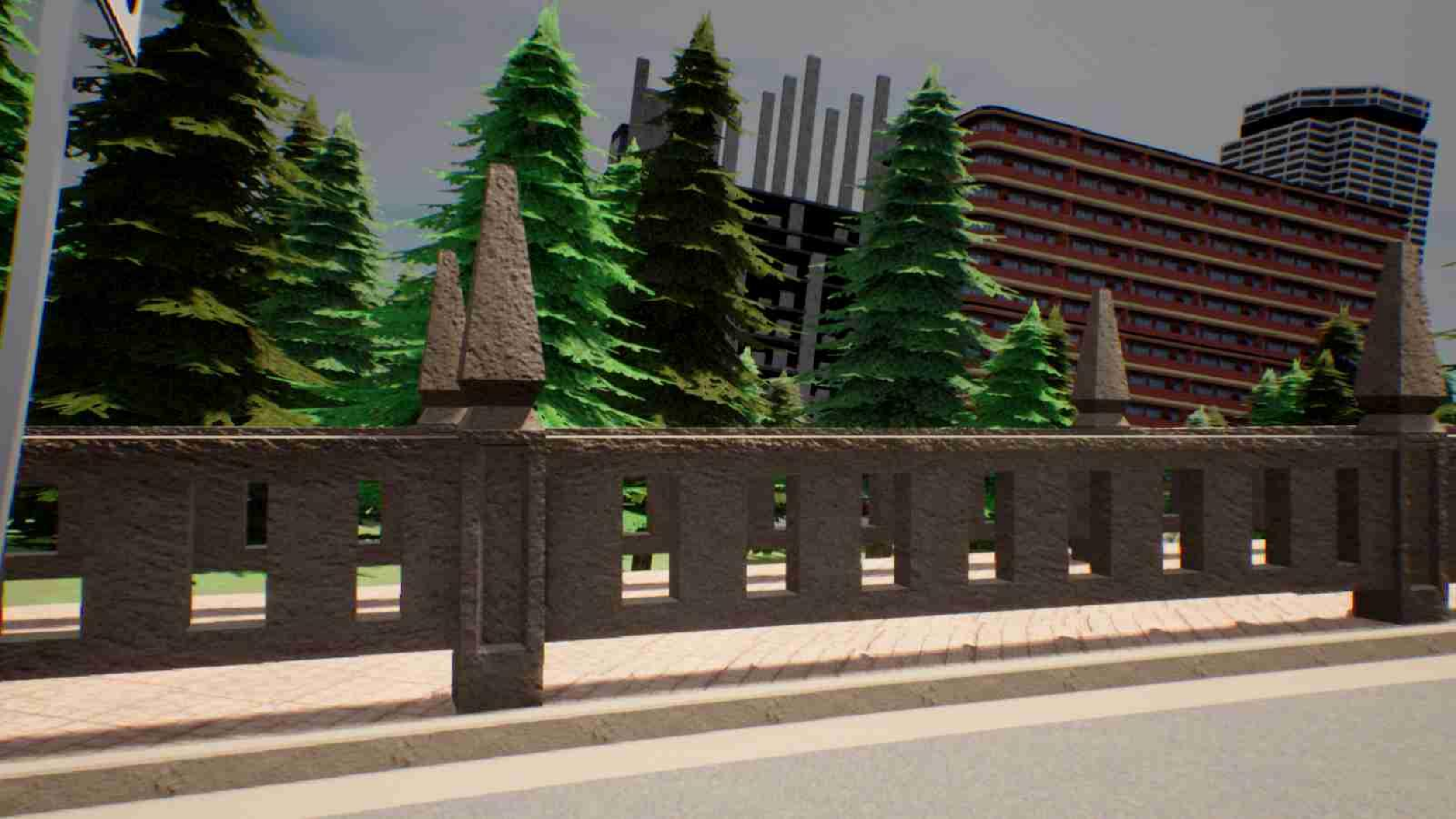}
\caption{Rear Right}
\end{subfigure}

\caption{Example observations from the six surround-view cameras used in the Orion system. 
The cameras provide a 360-degree perception coverage around the ego vehicle.}
\label{fig:orion_cameras}

\end{figure}

We follow the original sensor configurations used in the corresponding works to ensure fair comparison, as shown in \cref{tab:sensor_config}. Specifically, all sensor types, placements, and settings are kept consistent with the configurations reported in their original papers. An example of ORION camera settings are shown in \cref{fig:orion_cameras}.

\begin{table}[h]
\centering
\begin{tabular}{lcc}
\hline
\textbf{Method} & \textbf{Camera Configuration} & \textbf{LiDAR} \\
\hline
Orion & 6 surround-view cameras & -- \\
SimLingo & 1 front-view camera & -- \\
LMDrive & 4 cameras (multi-view) & 1 LiDAR \\
\hline
\end{tabular}
\caption{Sensor configurations of different autonomous driving methods.}
\label{tab:sensor_config}
\end{table}

\subsection{Method Implementation}

We compare our method with representative visual-language autonomous driving communication paradigm, including \textbf{visual-based communication} and \textbf{language-based communication} paradigms. Below we provide some detailed explanation for the implementation of the baseline paradigm.

\paragraph{Visual-based Communication.}
Different VLA models adopt different pipelines to process visual observations. Some models rely on a \textit{Q-Former} to convert visual features into language-aligned tokens (e.g., Orion and LMDrive), while others directly process images using a vision encoder (e.g., SimLingo).

Following the intermediate feature fusion paradigm, agents exchange visual tokens rather than raw images. Specifically, for Orion and LMDrive, we transmit the visual tokens produced by the Q-Former. For SimLingo, we transmit the visual tokens generated by the vision encoder.

During the final inference stage, the received visual tokens from collaborating agents are concatenated into the input token sequence of the VLA model. To distinguish the source of each visual observation, we prepend a prompt tag indicating the originating agent before the corresponding tokens.

\paragraph{Language-based Communication.}
Following prior work, we adopt a language-based collaboration paradigm where each agent first performs an autoregressive reasoning process similar to chain-of-thought (CoT). During this stage, the model analyzes the driving scene and produces structured reasoning about the environment.

The prompt used for this reasoning stage is shown below:

\begin{tcolorbox}[colback=gray!5,colframe=gray!50,title=Reasoning Prompt]
You are an autonomous driving assistant. Based on the current observation of the ego vehicle, analyze the driving scene step by step.

Provide the following:

\textbf{Object Description:} Describe important surrounding objects (type, position, motion).

\textbf{Scene Analysis:} Briefly analyze the road structure, traffic conditions, and potential risks.

\textbf{Intent Description:} Infer the possible intentions of nearby traffic participants.

\textbf{Action Reasoning:} Explain what the ego vehicle should do and why.

Keep the reasoning concise and structured.
\end{tcolorbox}

The generated reasoning results are collected and organized as structured textual information, which is then appended to the dialogue history and used as input to the VLA model for the final driving decision.

\paragraph{Our Method: LACO.}
In contrast to token-level communication, our method performs \textit{latent reasoning} during the observation prefill stage. This design allows the reasoning depth to be explicitly controlled without generating intermediate language tokens.

Instead of transmitting textual reasoning results, LACO selectively communicates the key-value (KV) cache representations in the latent space. At the final inference stage, the received latent KV states are directly incorporated into the reasoning process, avoiding redundant token-level recomputation. This design achieves a favorable balance between communication efficiency and decision performance.

\section{Further Experiment}
\begin{table}[h]
\centering
\small
\begin{tabular}{lcccccccc}
\toprule
 & \multicolumn{4}{c}{\textbf{Ours}} & \multicolumn{4}{c}{\textbf{Naive}} \\
\cmidrule(lr){2-5} \cmidrule(lr){6-9}
 & \multicolumn{2}{c}{V0} & \multicolumn{2}{c}{V1} 
 & \multicolumn{2}{c}{V0} & \multicolumn{2}{c}{V1} \\
\cmidrule(lr){2-3} \cmidrule(lr){4-5} \cmidrule(lr){6-7} \cmidrule(lr){8-9}
 & DS & RC & DS & RC & DS & RC & DS & RC \\
\midrule
ORION & 35.48 & 68.98 & 32.65 & 63.75 & 30.70 & 61.87 & 27.09 & 56.46 \\
SimLingo & 35.73 & 72.06 & 29.22 & 68.00 & 28.36 & 54.73 & 23.48 & 41.11 \\
LMDrive (LLaMA) & 22.84 & 39.55 & 30.54 & 61.34 & 19.08 & 40.26 & 25.41 & 52.05 \\
LMDrive (LLaVA) & 28.40 & 48.93 & 30.13 & 52.96 & 23.63 & 34.58 & 24.72 & 40.52 \\
LMDrive (Vicuna) & 32.07 & 61.36 & 31.41 & 51.56 & 26.13 & 51.44 & 25.85 & 38.71 \\
\bottomrule
\end{tabular}
\caption{Comparison between LACO and naive latent communication strategy. }
\label{tab:naive_compare}
\end{table}
\paragraph{Result Analysis.}

\cref{tab:naive_compare} reports the comparison between LACO and the naive latent KV communication strategy across multiple VLA backbones. 
Overall, LACO consistently achieves higher Driving Score (DS) and Route Completion (RC) than the naive strategy in most settings.

For the vision-language models ORION and SimLingo, the advantage of LACO is particularly clear. 
For example, on ORION (V0), LACO improves DS from 30.70 to 35.48 and RC from 61.87 to 68.98. 
Similarly, on SimLingo (V0), DS increases from 28.36 to 35.73 while RC improves from 54.73 to 72.06. 
These results indicate that selectively transmitting informative latent representations allows agents to better leverage collaborative information while avoiding unnecessary interference.

For LMDrive variants, LACO also demonstrates consistent improvements in most configurations. 
Although the naive strategy occasionally achieves comparable performance in certain settings, it remains less stable across different model backbones and scenario splits. 
This variability suggests that directly sharing the full latent KV cache introduces noisy intermediate reasoning states that may interfere with downstream attention.

\paragraph{Conclusion.}

Overall, the results demonstrate that simply sharing the full latent KV cache is insufficient for effective multi-agent collaboration. 
In contrast, LACO enables more reliable and efficient collaboration by selectively transmitting informative latent representations, which mitigates agent information confusion and improves decision quality across diverse VLA architectures.

\end{document}